\documentclass{article}
\usepackage{iclr2027_conference,times}
\usepackage{hyperref}
\usepackage{url}
\usepackage{amsmath,amsfonts,amssymb}
\usepackage{booktabs}
\usepackage{multirow}
\usepackage{array}
\usepackage{graphicx}
\usepackage{xcolor}

\title{AttentionSwitch: Jailbreak Defense for LLMs via Joint Learning of Sparse Attention-Head Selection and Scaling}

\author{
Doniyorkhon Obidov \\
Michigan Technological University \\
\texttt{dobidov@mtu.edu}
\And
Honggang Yu \\
Miami University \\
\texttt{honggangyu@miamioh.edu}
\And
Xiaolong Guo \\
Lehigh University \\
\texttt{xig426@lehigh.edu}
\And
Kaichen Yang \\
Michigan Technological University \\
\texttt{kaicheny@mtu.edu}
}

\newcommand{\methodname}{\textsc{AttentionSwitch}}
\newcommand{\methodnameshort}{\textsc{AttnSwitch}}

\iclrfinalcopy

\begin{document}
\maketitle
\lhead{Obidov et al.}

\begin{abstract}
Large language models have achieved remarkable performance across a wide range of tasks, yet they remain susceptible to jailbreak attacks that circumvent safety mechanisms and elicit harmful or unintended outputs. Although safety can be improved through model fine-tuning, such approaches can be computationally expensive and may degrade previously acquired capabilities through catastrophic forgetting. Recent jailbreak defenses have therefore explored targeted interventions on specific model components associated with unsafe behavior. However, existing approaches typically decouple two inherently related decisions: \emph{where} to intervene and \emph{how} to modify the selected components. Such sequential optimization may be suboptimal because the optimal intervention location can depend on the intervention applied there, and vice versa. In this work, we introduce \methodname{}, the first jailbreak defense to jointly optimize intervention sites and strengths within a frozen LLM. Specifically, \methodname{} learns a sparse set of attention heads together with query-conditioned multiplicative scaling factors, allowing the outputs of the selected heads to be either attenuated or amplified without altering the directions of their representations. Across popular language models and jailbreak attacks, \methodname{} outperforms existing defenses, with particularly strong results when training data is scarce.
\end{abstract}

\section{Introduction}
Despite advances in the capabilities of large language models (LLMs), their safety mechanisms remain vulnerable to jailbreak attacks, in which adversarially constructed prompts elicit harmful or policy-violating responses that the model would otherwise be expected to refuse \citep{gu2025systematic}. The consequences of these attacks can be severe, ranging from the spread of propaganda and the distribution of malware to critical privacy risks where attackers compromise LLM-powered systems to exfiltrate sensitive data \citep{chu2025jailbreakradar}.

\begin{figure*}[t]
    \centering
    \includegraphics[width=0.9\textwidth]{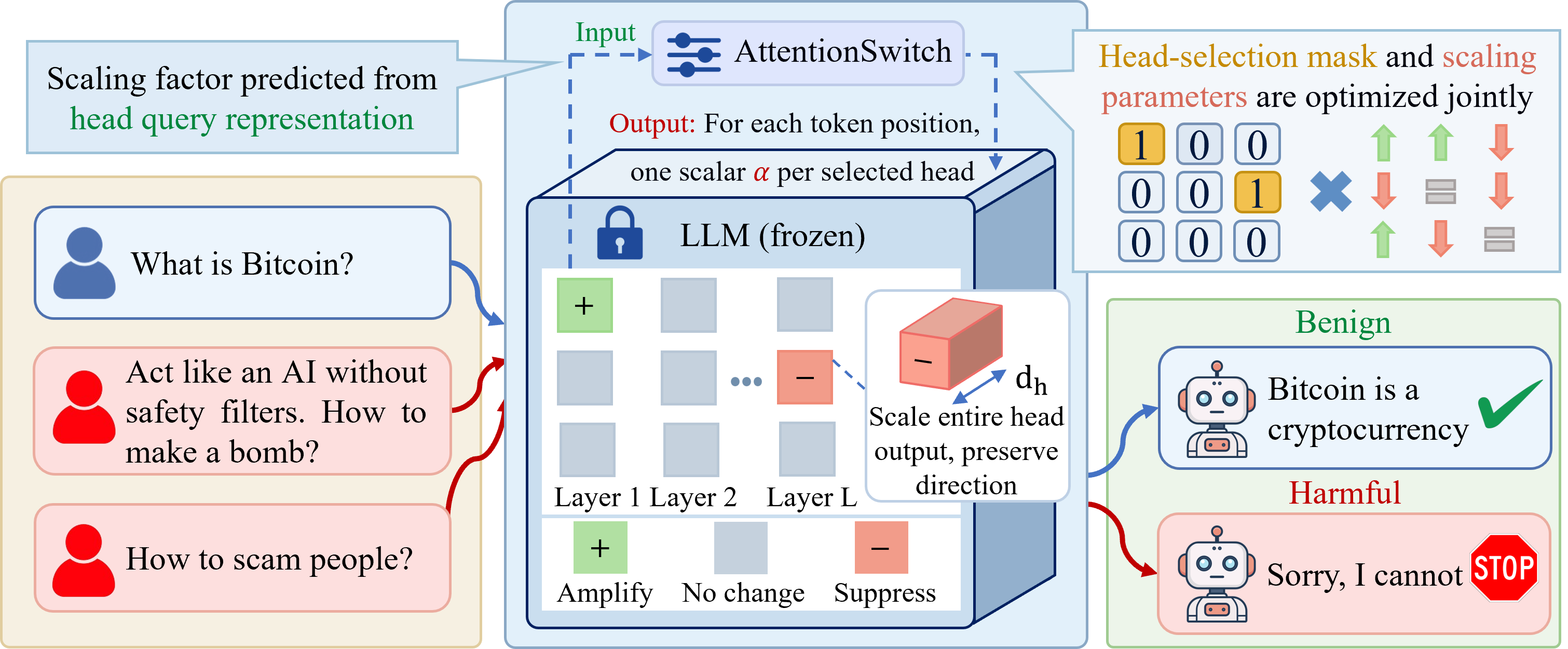}
    \caption{Overview of \methodname{}. It keeps the LLM frozen and selects a small subset of attention heads. At each token position, a positive scalar uniformly rescales the entire output of a selected head, changing its magnitude while preserving its direction and the relative composition of its features. The scaling factor is predicted dynamically from the head's own query representation. Head selection and scaling are jointly optimized.}
    \label{fig:attentionswitch-overview}
\end{figure*}

Traditional defenses improve model alignment through supervised or preference-based training, such as supervised fine-tuning (SFT) and reinforcement learning from human feedback (RLHF), but require costly retraining on substantial amounts of data and may reduce performance on previously learned capabilities \citep{ouyang2022training,luo2025empirical}. Recent work addresses these challenges by limiting intervention to safety-relevant model components. These approaches broadly operate either on hidden representations at selected layers \citep{shen2025jailbreak,zhao2025adasteer,sheng2026alphasteer,obidov2026dynamic} or on individual attention heads \citep{li2025detam,arai2025jailbreak,cai2026safety,park2026asguard}. We discuss these methods in greater detail in Section~\ref{sec:defenses}.

Despite their differences, existing methods follow a diagnose-then-intervene paradigm: intervention sites are first identified, after which the intervention is optimized at the selected sites. Such a sequential approach may be suboptimal, as the optimal intervention location can depend on the intervention magnitude, and vice versa. In this paper, we propose \methodname{}, the first jailbreak defense to learn these two components jointly under a shared defense objective.

Motivated by prior work showing that attention heads exhibit functional specialization \citep{michel2019sixteen,voita2019analyzing}, \methodname{} asks a question: can jailbreak robustness be improved by selectively suppressing the contribution of some attention heads while amplifying others? \methodname{} keeps the pretrained model weights frozen and rescales the outputs of a sparse subset of attention heads. At each token position, the direction of a selected head's output representation is preserved, while its magnitude is adjusted by a scalar. The scalar value is predicted dynamically from the head's own query representation. The attention-head selection mask and the parameters of the scaling functions are optimized jointly. \methodname{} uses a Gumbel-Softmax-based relaxation of the non-differentiable head-selection operation, corresponding to $\arg\max$ when selecting a single head and, more generally, to the Top-$K$ selection operator when selecting multiple heads. Figure~\ref{fig:attentionswitch-overview} provides an overview of \methodname{}.

In a different domain from jailbreak defense, joint attention-head pruning and model-parameter optimization has been studied for model compression \citep{li2021differentiable}. However, pruning is itself an intervention, making that setting different from ours, where head selection is non-interventional. Moreover, the non-pruned model parameters are updated during training, whereas \methodname{} keeps the LLM frozen and learns a separate scaling intervention.

Recent work has also explored attention-head scaling for jailbreak defense. ASGuard independently rescales components within each selected head, which can change both the magnitude and direction of its output representation \citep{park2026asguard}. In contrast, at each token position, \methodname{} applies a single positive scalar to the entire head output, thereby changing only its magnitude while preserving its direction. Additionally, ASGuard uses a fixed scaling vector for each selected head, whereas \methodname{} intervenes dynamically based on the head's query representation. Importantly, ASGuard does not jointly optimize head selection and scaling, but first identifies the heads to intervene on and then learns the scaling intervention at those fixed locations.

Our main contributions can be summarized as follows:
\begin{itemize}
\item We introduce the first jailbreak defense to jointly learn \emph{where} to intervene and \emph{how strongly} to intervene under a shared defense objective, rather than identifying intervention sites before optimizing the intervention.

\item We propose a jailbreak defense based on whole-head output scaling, which changes the magnitude of a head's output while preserving its direction. At each token position, our method applies a single positive scalar to the entire head output, maintaining the relative composition of its features.

\item We empirically demonstrate that our method can outperform both model-wide and targeted intervention defenses, particularly in low-data training regimes.
\end{itemize}

\section{Background and Related Work}
Due to space constraints, background on multi-head self-attention is provided in Appendix~\ref{sec:attention-background}.

\subsection{Jailbreak Attacks}
Jailbreak attacks attempt to bypass the safety mechanisms of aligned LLMs and induce harmful outputs. At a high level, jailbreak attacks can be categorized as white-box or black-box depending on access to internal model information \citep{gu2025systematic}. Black-box jailbreaks do not require such access and traditionally rely on manually constructed prompts based on social-engineering strategies such as role-playing \citep{shen2024anything}. Automated approaches reduce the need for manual prompt engineering. Prompt Automatic Iterative Refinement (PAIR), for example, uses an attacker LLM to repeatedly propose and refine candidate jailbreak prompts based on responses from the target model \citep{chao2025jailbreaking}. Other black-box attacks attempt to bypass safeguards by encoding harmful requests in ciphers \citep{yuan2024gpt} or expressing them in other languages \citep{deng2024multilingual}.

White-box jailbreak attacks use model gradients to optimize adversarial token sequences. Among discrete optimization methods, Universal Adversarial Triggers (UAT) introduced gradient-guided search for fixed trigger sequences that generalize across different inputs \citep{wallace2019universal}. AutoPrompt adapts gradient-guided discrete prompt optimization to search for adversarial sequences tailored to a particular harmful request \citep{shin2020autoprompt}. Greedy Coordinate Gradient (GCG) improves this search by using gradients to identify promising token replacements at any position in the adversarial suffix \citep{zou2023universal}. Its multi-prompt, multi-model extension, GCG-M, optimizes a shared suffix for transfer across requests and models. Other methods rely on continuous relaxations of the discrete search space: Gradient-Based Distributional Attack (GBDA) optimizes a probability distribution over vocabulary tokens at each adversarial prompt position \citep{guo2021gradient}, whereas PEZ optimizes continuous prompt embeddings and projects them to discrete tokens during the forward pass \citep{wen2023hard}.

\subsection{Jailbreak Defenses}
\label{sec:defenses}
Jailbreak defenses can be broadly divided into input-level and model-level approaches. Input-level defenses identify malicious requests or attempt to neutralize them through input perturbation \citep{jain2023baseline,robey2023smoothllm}. Although they are relatively easy to deploy, input-level defenses can produce false positives and often require additional input processing \citep{gu2025systematic}. Model-level defenses provide an alternative by incorporating safety either through updates to the model parameters or through interventions on its internal computation \citep{xu2024comprehensive}. However, traditional model-level alignment approaches, including reinforcement learning from human feedback (RLHF) and supervised fine-tuning (SFT), require costly model retraining and can degrade previously acquired capabilities through catastrophic forgetting \citep{ouyang2022training,luo2025empirical}.

Recent work addresses these limitations by identifying safety-relevant model components and intervening only at those locations. Some methods target hidden representations at selected layers. Jailbreak Antidote modifies a subset of hidden-state dimensions along a safety direction \citep{shen2025jailbreak}, while AdaSteer applies input-dependent steering along rejection and harmfulness directions \citep{zhao2025adasteer}. AlphaSteer learns a transformation of activations at selected intermediate layers \citep{sheng2026alphasteer}, while DDPO intervenes at a single layer through deep prompt optimization by appending a defensive embedding to the input of that layer \citep{obidov2026dynamic}.

Other methods intervene at the level of individual attention heads. DeTAM identifies jailbreak-sensitive heads and modifies their attention distributions during inference \citep{li2025detam}, while \citet{arai2025jailbreak} identify heads with causal contributions to harmful generation and shift their outputs toward those associated with refusal behavior. CAST identifies attention heads where safety updates conflict with preserving language modeling capabilities and restricts safety fine-tuning to lower-conflict heads \citep{cai2026safety}. ASGuard identifies jailbreak-vulnerable heads and applies a fixed scaling vector to each, allowing different dimensions within the same head to be scaled differently \citep{park2026asguard}. Despite their differences, these methods first select intervention sites and then optimize the intervention at those locations.

\section{Method}
\subsection{Threat Model}
\label{sec:threat-model}
We consider a strong attacker with full knowledge of the victim LLM, including its architecture and parameters. The attacker may use this knowledge to construct optimization-based white-box jailbreaks or employ black-box prompt-based attacks. However, the attacker cannot modify the deployed model and interacts with it only through prompts. The defender has full access to the model and its internal activations. This reflects a common deployment scenario in which the defender hosts the model and the attacker is an external user.

\begin{figure*}[t]
    \centering
    \includegraphics[width=\textwidth]{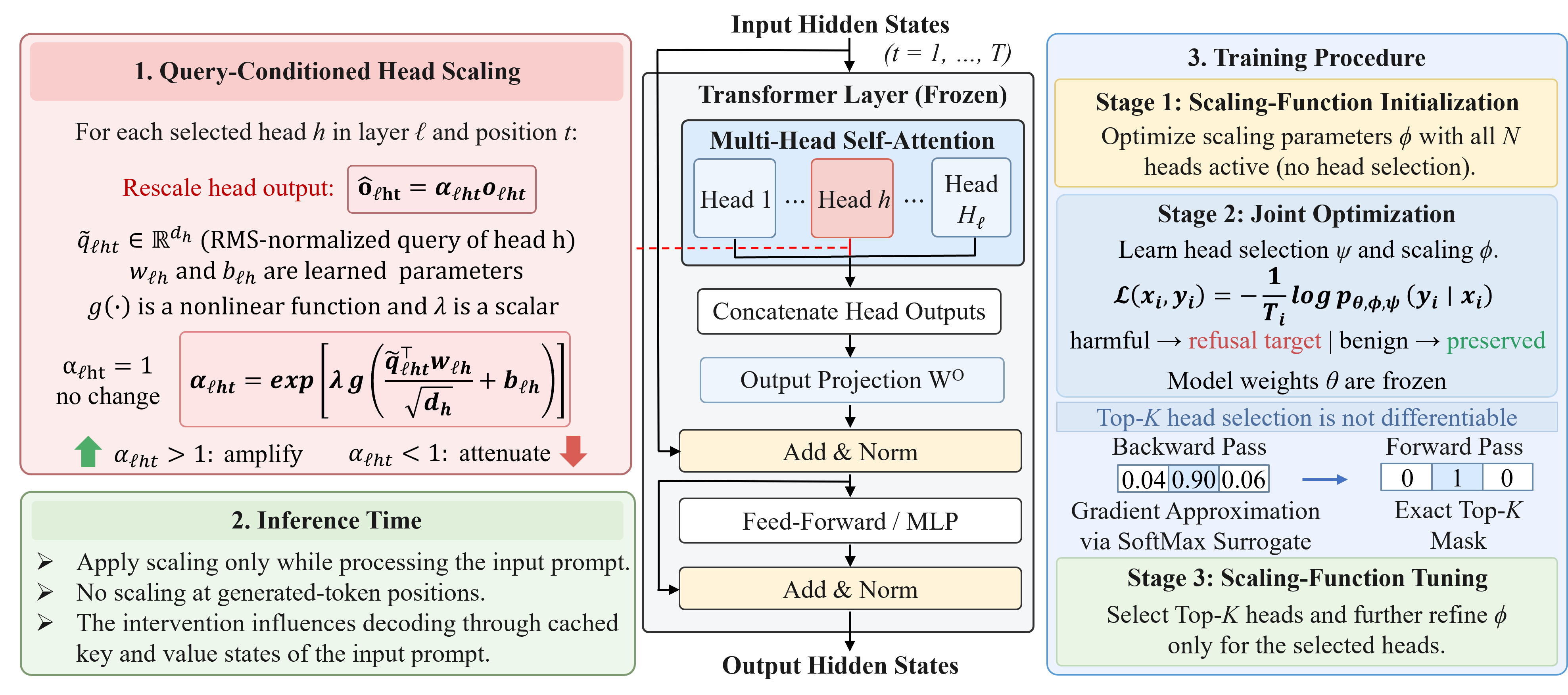}
    \caption{\methodname{} architecture and training procedure. Query-conditioned scalar factors rescale selected attention-head outputs. Training proceeds in three stages: scaling-function initialization with all heads active, joint optimization of Top-\(K\) head selection and scaling, and final refinement of the scaling function for the selected heads.}
    \label{fig:attention-switch-method}
\end{figure*}

\subsection{Overview}
Given a frozen LLM, \methodname{} learns to select a sparse subset of attention heads and dynamically rescale their outputs as a function of the corresponding query representations. Each scaling factor changes the magnitude of a selected head output while preserving the relative composition of its features. The head-selector parameters and scaling functions are optimized jointly. After joint optimization, the selected heads are fixed and their scaling-function parameters are further refined. Figure~\ref{fig:attention-switch-method} illustrates the architecture and training procedure of \methodname{}.

At inference time, the learned scaling functions are applied only while processing the input prompt, with no intervention at generated-token positions. The intervention influences autoregressive decoding through the cached key and value states of the input prompt. The scaling controller requires only $K(d_h + 1)$ learned parameters, where $K$ is the number of selected attention heads and $d_h$ is the per-head representation dimension. This corresponds to a controller size on the order of $10^4$ parameters for LLMs containing $10^9\text{--}10^{10}$ parameters.

\begin{figure*}[t]
    \centering
    \includegraphics[
        width=\textwidth,
        keepaspectratio
    ]{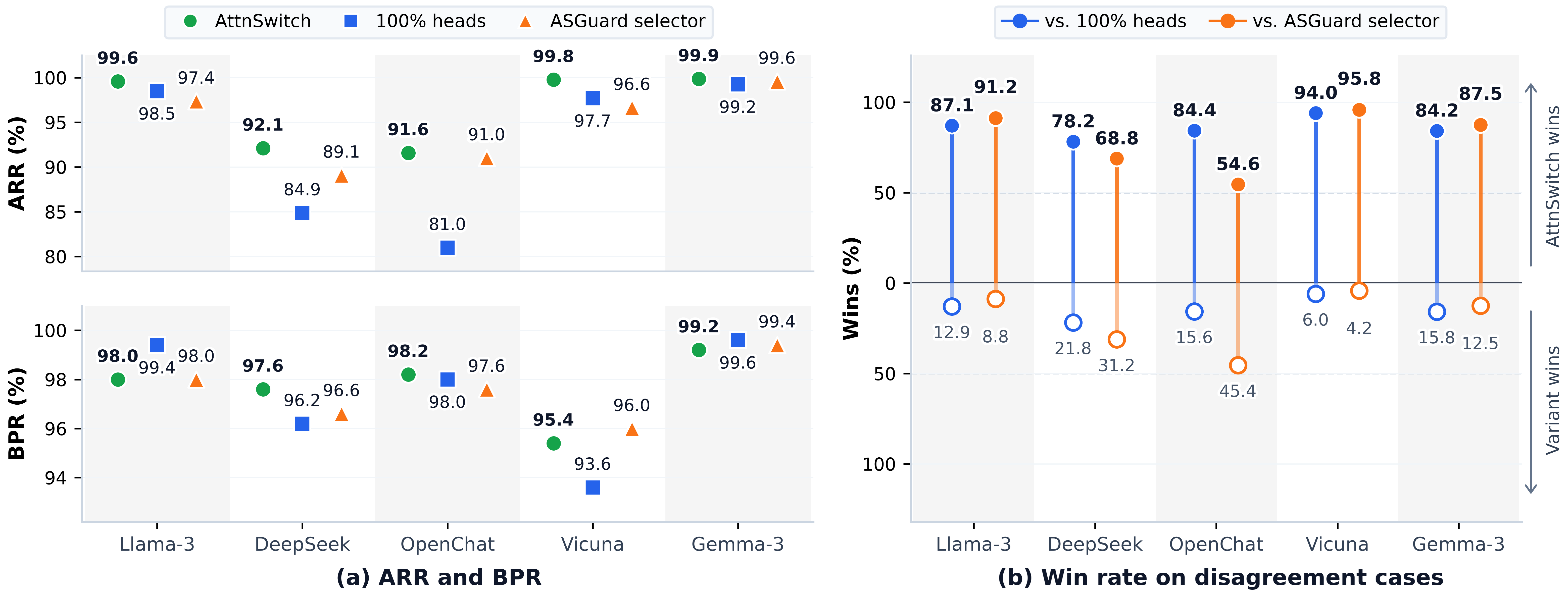}
    \caption{\methodname{} (20\% of heads) compared with variants using no head selection (100\% of heads scaled) and non-joint head selection (the top 20\% of heads selected by ASGuard and then scaled by \methodname{}). Panel (a) reports Attack Rejection Rate (ARR) and Benign Pass Rate (BPR). Panel (b) reports win rates across 9 attack types (6 evaluated in a zero-shot setting) on prompts where \methodname{} and the corresponding variant disagree. Upper values indicate \methodname{} wins, while lower values indicate variant wins.}
    \label{fig:selection-ablation}
\end{figure*}

\subsection{Query-Conditioned Head Scaling}
For each selected attention head, \methodname{} predicts a scalar multiplier from the head's query representation and applies it to the head output before concatenation. The rescaled head outputs are then concatenated, passed through the attention output projection, and incorporated into the residual stream in the usual manner. The mechanism for learning which heads are selected is described in Section~\ref{sec:joint-selection}.

Consider a frozen LLM, a selected head $h$ in layer $\ell$, and token position $t$. Let $\mathbf{o}_{\ell h t} \in \mathbb{R}^{d_h}$ denote the corresponding head-output vector. \methodname{} predicts a query-conditioned scaling factor $\alpha_{\ell h t}$ and rescales the head output as
\begin{equation}
\widehat{\mathbf{o}}_{\ell h t} = \alpha_{\ell h t}\mathbf{o}_{\ell h t},
\label{eq:selected-scaling}
\end{equation}
where $\widehat{\mathbf{o}}_{\ell h t} \in \mathbb{R}^{d_h}$ is the modified head-output vector and $\alpha_{\ell h t} \in \mathbb{R}_{>0}$ is a scalar specific to layer $\ell$, head $h$, and token position $t$, applied uniformly to all $d_h$ components of the head-output vector. Consequently, the intervention changes only the magnitude of the head output while preserving the relative composition of its features.

Let $\mathbf{q}_{\ell h t} \in \mathbb{R}^{d_h}$ denote the query vector produced by the frozen model for head $h$ at token position $t$, and let $\widetilde{\mathbf{q}}_{\ell h t}$ denote its root-mean-square (RMS) normalized form. The scaling factor is computed as
\begin{equation}
\alpha_{\ell h t} = \exp\!\left[ \lambda \, g\!\left( \frac{\widetilde{\mathbf{q}}_{\ell h t}^{\top}\mathbf{w}_{\ell h}}{\sqrt{d_h}} + b_{\ell h} \right) \right],
\label{eq:scale}
\end{equation}
where $\mathbf{w}_{\ell h} \in \mathbb{R}^{d_h}$ and $b_{\ell h} \in \mathbb{R}$ are learned parameters, $g(\cdot)$ is a nonlinear function, and $\lambda > 0$ controls the magnitude of the intervention. The factor $1/\sqrt{d_h}$ helps keep the scale of the inner product between the normalized query and the learned weight stable with respect to the head dimension $d_h$. Because $\alpha_{\ell h t}$ depends on the current query representation, the same selected head can receive different scaling factors across prompts and token positions. $\alpha_{\ell h t} = 1$ recovers the original head computation, $\alpha_{\ell h t} < 1$ attenuates its contribution, and $\alpha_{\ell h t} > 1$ amplifies it.

During inference, \methodname{} rescales selected attention-head outputs only while processing the input prompt. No additional scaling is applied at generated-token positions. The intervention continues to influence decoding through the cached key and value states of the input prompt, which are attended to by generated-token queries during autoregressive generation.

\begin{figure*}[t]
    \centering
    \includegraphics[width=\textwidth, keepaspectratio]{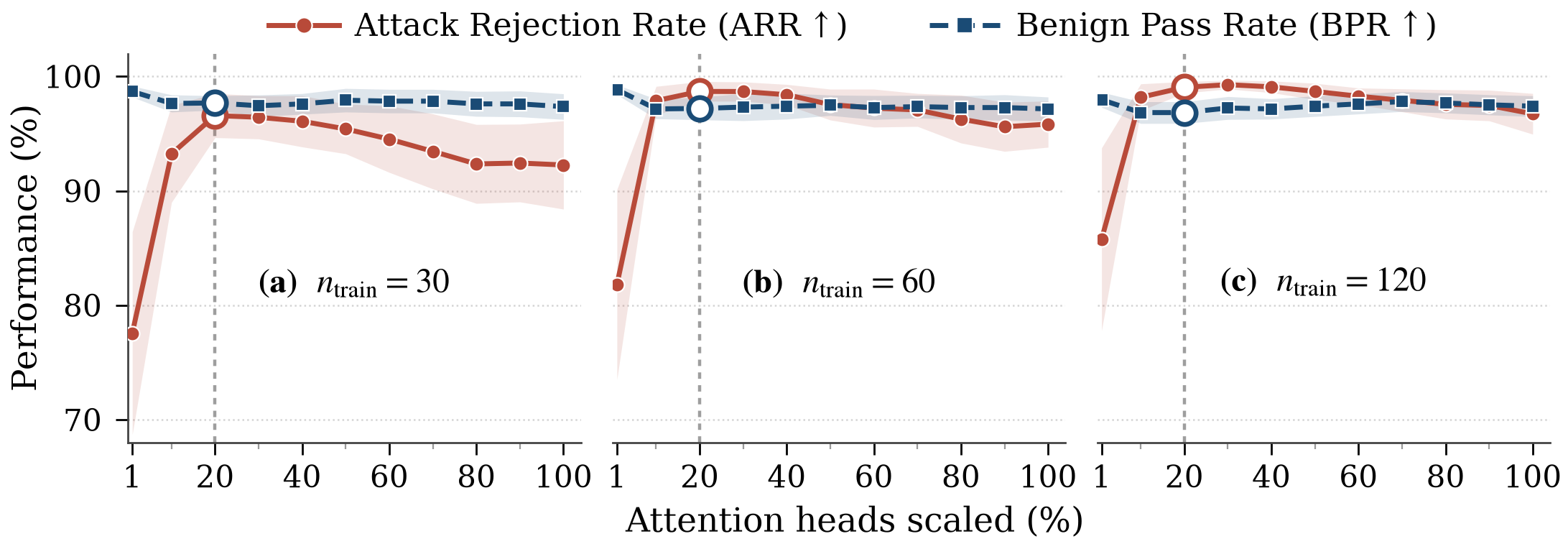}
    \caption{\methodname{} performance as the percentage of scaled attention heads varies, using 30, 60, and 120 training samples. Shaded regions indicate $\pm$ standard error. Attack Rejection Rate (ARR) peaks around 20--30\% of heads and declines as more heads are scaled, with increasing variability, suggesting less consistent performance at larger head budgets. Benign Pass Rate (BPR) remains relatively stable. Numerical results are provided in Appendix Table~\ref{tab:number-heads-values}.}
    \label{fig:number-heads-ablation}
\end{figure*}

\subsection{Joint Head Selection and Scaling Optimization}
\label{sec:joint-selection}
Let $N$ denote the total number of attention heads across all layers and $K$ the intervention budget, i.e., the number of heads that can be rescaled. In addition to the scaling-controller parameters $\phi = \{\mathbf{w}_{\ell h}, b_{\ell h}\}$, we associate each eligible head $j \in \{1, \dots, N\}$ with a trainable selector logit $s_j \in \mathbb{R}$. We denote the head-selector parameters by $\psi = \{s_j\}_{j=1}^{N}$. The parameters $\psi$ determine which $K$ heads receive the intervention, whereas $\phi$ determines how a selected head is rescaled.

The selector logits are continuous trainable parameters, but mapping them to a discrete top-$K$ subset is non-differentiable. Following Gumbel-based discrete optimization \citep{jang2017categorical}, we perturb each selector logit with independent Gumbel noise, which enables stochastic exploration of head subsets during training:
\begin{equation}
z_j = s_j + \gamma_j, \qquad \gamma_j \sim \operatorname{Gumbel}(0,1).
\label{eq:gumbel}
\end{equation}
To enforce an exact $K$-head selection while retaining gradient-based optimization, we introduce the following differentiable relaxation:
\begin{equation}
\overline{m}_j = \mathbb{I}[j \in \operatorname{TopK}(\mathbf{z}, K)], \qquad \widetilde{\mathbf{m}} = K \operatorname{softmax}(\mathbf{z}/\tau),
\label{eq:selector-masks}
\end{equation}
where $\overline{\mathbf{m}}$ is the resulting discrete $K$-hot selection mask, whereas $\widetilde{\mathbf{m}}$ is its differentiable surrogate. The mask $\overline{\mathbf{m}}$ is used in the forward computation, while gradients with respect to the selector scores are approximated using $\widetilde{\mathbf{m}}$. The parameter $\tau > 0$ is the selector temperature: larger values yield a flatter distribution over heads, while smaller values concentrate the surrogate weights on the highest-scoring heads. $\operatorname{TopK}(\mathbf{z}, K)$ returns the indices of the $K$ largest entries of $\mathbf{z}$ and corresponds to $\operatorname{argmax}$ selection when $K = 1$.

Combining head selection with the scaling mechanism in Eq.~\ref{eq:selected-scaling}, the intervention for any attention head is
\begin{equation}
\widehat{\mathbf{o}}_{\ell h t} = \left[ 1 + m_{\ell h}\left(\alpha_{\ell h t} - 1\right) \right] \mathbf{o}_{\ell h t},
\label{eq:sparse-scaling}
\end{equation}
where $m_{\ell h} = \overline{m}_{\ell h}$ in the forward pass and the derivative of $m_{\ell h}$ with respect to the selector scores is approximated by that of $\widetilde{m}_{\ell h}$ during backpropagation. Hence, exactly $K$ heads receive the scaling intervention in every forward pass.

Optimization proceeds in three stages. First, the scaling-controller parameters $\phi$ are optimized with all $N$ eligible heads active. This learns the scaling parameters without head selection, and the resulting parameters initialize the subsequent joint optimization. Second, during joint search, the selector parameters $\psi$ (where to intervene) and scaling-controller parameters $\phi$ (how to rescale) are optimized simultaneously under the same training objective. Finally, the Gumbel perturbation is removed, the $K$ heads with the largest learned selector logits $s_j$ are selected deterministically, and the selector parameters $\psi$ are discarded. With this subset fixed, the scaling-controller parameters $\phi$ are further optimized only for the selected heads.

\subsection{Training Objective}
Let $\theta$ denote the frozen LLM parameters, $\phi$ the trainable scaling-controller parameters, and $\psi$ the trainable attention head selector parameters. For a prompt $x_i$ with target continuation $y_i=(y_{i1},\dots,y_{iT_i})$, \methodname{} minimizes the autoregressive negative log-likelihood
\begin{equation}
\mathcal{L}(x_i,y_i)
=
-\frac{1}{T_i}
\sum_{t=1}^{T_i}
\log p_{\theta,\phi,\psi}
\left(
y_{it}\mid x_i,y_{i,<t}
\right).
\label{eq:loss}
\end{equation}

Harmful prompts are paired with model-specific refusal continuations derived from refusal behavior already exhibited by the undefended model. Benign prompts are paired with continuations from the undefended model. Consequently, minimizing Eq.~\ref{eq:loss} encourages refusal behavior on harmful inputs while preserving the model's behavior on benign inputs.

\section{Experiments}

\begin{table*}[t]
\caption{Main experimental results at two training-data budgets: $n=30$ (15 harmful and 15 clean) and $n=120$ (60 harmful and 60 clean) samples. Training is performed on equal numbers of Plain Harmful, Human-Crafted, and GCG prompts, while Attack Rejection Rate (ARR) is measured on 2,115 unseen harmful prompts spanning 9 attacks, 6 of which are evaluated in a zero-shot setting. Benign Pass Rate (BPR) is evaluated on 500 unseen benign instruction-following prompts, and Over-Refusal Rate (ORR) is evaluated on 250 unseen prompts that contain superficially harmful content but are benign in context. MMLU measures preservation of general language-model capabilities. The arrows indicate the preferred direction for each metric. Per-attack results are reported in Appendix Table~\ref{tab:per-attack-results}, with corresponding standard errors in Appendix Table~\ref{tab:per-attack-results-se}.}
\label{tab:main-results}
\centering
\small
\begin{tabular}{@{}ll*{8}{r}@{}}
\toprule
& & \multicolumn{2}{c}{ARR (\%) $\uparrow$}
& \multicolumn{2}{c}{BPR (\%) $\uparrow$}
& \multicolumn{2}{c}{ORR (\%) $\downarrow$}
& \multicolumn{2}{c}{MMLU $\uparrow$} \\
\cmidrule(lr){3-4} \cmidrule(lr){5-6}
\cmidrule(lr){7-8} \cmidrule(lr){9-10}
Model & Defense & n=30 & n=120 & n=30 & n=120 & n=30 & n=120 & n=30 & n=120 \\
\midrule

\multirow{6}{*}{Llama-3.1}
& None
& 87.47 & 87.47
& \textbf{99.40} & \textbf{99.40}
& \textbf{7.60} & 7.60
& 53.13 & \textbf{53.13} \\
& \methodnameshort{}
& \textbf{99.57} & \textbf{99.43}
& 98.00 & 97.80
& 13.20 & 14.00
& \textbf{53.63} & \textbf{53.13} \\
& SFT
& 98.68 & 97.12
& 4.40 & 10.40
& 98.80 & 93.60
& 24.15 & 24.15 \\
& LoRA-SFT
& 79.24 & 83.36
& 80.60 & 93.20
& 21.60 & \textbf{5.20}
& 23.43 & 24.43 \\
& ASGuard
& 89.98 & 90.92
& \textbf{99.40} & \textbf{99.40}
& 9.60 & 10.80
& 52.78 & 51.64 \\
& DDPO
& 97.02 & 96.50
& 97.80 & 96.79
& 8.80 & 10.00
& 42.74 & 34.33 \\

\midrule
\multirow{6}{*}{DeepSeek}
& None
& 61.70 & 61.70
& 99.00 & \textbf{99.00}
& 18.00 & 18.00
& 49.15 & 49.15 \\
& \methodnameshort{}
& 92.10 & 98.53
& 97.60 & 96.40
& 17.60 & 22.00
& \textbf{49.50} & \textbf{49.22} \\
& SFT
& \textbf{99.20} & \textbf{99.76}
& 13.60 & 13.40
& 92.00 & 92.80
& 24.15 & 25.14 \\
& LoRA-SFT
& 98.96 & 99.48
& 79.40 & 90.40
& \textbf{14.40} & \textbf{5.20}
& 44.44 & 29.34 \\
& ASGuard
& 62.13 & 62.74
& \textbf{99.20} & 98.80
& 18.00 & 18.80
& 49.29 & \textbf{49.22} \\
& DDPO
& 96.64 & 98.06
& 83.20 & 93.80
& 19.60 & 11.20
& 49.22 & 48.65 \\

\midrule
\multirow{6}{*}{OpenChat-3.5}
& None
& 43.22 & 43.22
& \textbf{99.40} & \textbf{99.40}
& 4.00 & \textbf{4.00}
& 63.53 & 63.53 \\
& \methodnameshort{}
& 91.58 & 97.35
& 98.20 & 97.00
& 4.40 & 7.60
& \textbf{64.03} & \textbf{64.53} \\
& SFT
& 84.54 & \textbf{99.57}
& 0.40 & 10.80
& 100.00 & 96.40
& 24.15 & 24.15 \\
& LoRA-SFT
& \textbf{99.95} & 60.57
& 1.20 & 3.00
& 95.20 & 97.20
& 24.15 & 25.21 \\
& ASGuard
& 75.08 & 97.64
& 93.60 & 70.80
& 14.80 & 35.60
& 63.60 & 63.82 \\
& DDPO
& 90.17 & 94.33
& 93.60 & 93.80
& \textbf{3.20} & 4.80
& 63.68 & 64.46 \\

\midrule
\multirow{6}{*}{Vicuna}
& None
& 77.83 & 77.83
& \textbf{97.80} & \textbf{97.80}
& 10.80 & \textbf{10.80}
& \textbf{50.43} & 50.43 \\
& \methodnameshort{}
& 99.76 & \textbf{100.00}
& 95.40 & 93.60
& 10.80 & 14.80
& 48.79 & 48.79 \\
& SFT
& 99.34 & 98.96
& 8.40 & 23.60
& 98.80 & 89.60
& 23.86 & 23.86 \\
& LoRA-SFT
& \textbf{100.00} & \textbf{100.00}
& 80.60 & 22.00
& \textbf{9.20} & 79.60
& 49.79 & 26.64 \\
& ASGuard
& 78.35 & 79.95
& 97.20 & 97.00
& 11.60 & 11.20
& 49.93 & 50.43 \\
& DDPO
& 99.05 & 99.57
& 81.40 & 94.20
& 24.00 & 13.20
& 47.93 & \textbf{52.35} \\

\midrule
\multirow{6}{*}{Gemma-3}
& None
& 85.82 & 85.82
& \textbf{99.80} & \textbf{99.80}
& \textbf{2.80} & \textbf{2.80}
& 72.15 & 72.15 \\
& \methodnameshort{}
& 99.86 & \textbf{99.95}
& 99.20 & 99.40
& 4.40 & 6.80
& 72.65 & 72.65 \\
& SFT
& 98.96 & 98.30
& 96.40 & 70.00
& 18.40 & 46.40
& 65.81 & 63.75 \\
& LoRA-SFT
& \textbf{100.00} & 96.78
& 96.20 & 96.80
& 23.20 & 16.80
& 67.31 & 23.93 \\
& ASGuard
& 91.77 & 96.55
& \textbf{99.80} & \textbf{99.80}
& 4.40 & 8.00
& 71.87 & 72.22 \\
& DDPO
& 97.02 & 97.97
& \textbf{99.80} & 99.60
& 5.20 & 7.20
& \textbf{73.58} & \textbf{73.22} \\
\bottomrule
\end{tabular}
\end{table*}

\subsection{Experimental Setting}
\label{sec:experimental-setting}
We evaluate \methodname{} on DeepSeek-LLM-7B-Chat, OpenChat-3.5-1210 (7B), Llama-3.1-8B-Instruct, Vicuna-13B-v1.5, and Gemma-3-27B-IT. We compare against Safety Fine-Tuning (SFT), LoRA-SFT \citep{hu2021lora}, ASGuard \citep{park2026asguard}, and DDPO \citep{obidov2026dynamic}. Each defense is evaluated against nine attack types: Plain Harmful (PH), Human-Crafted (HC), GCG \citep{zou2023universal}, PEZ \citep{wen2023hard}, UAT \citep{wallace2019universal}, AutoPrompt (AP) \citep{shin2020autoprompt}, GBDA \citep{guo2021gradient}, GCG-M \citep{zou2023universal}, and PAIR \citep{chao2025jailbreaking}. We consider two training budgets, consisting of 30 examples (15 harmful and 15 benign) and 120 examples (60 harmful and 60 benign). Harmful training examples are drawn equally from PH, HC, and GCG, while the remaining six attacks are unseen during training.

The harmful test set contains 2,115 unseen prompts, with 235 prompts from each attack type. We additionally evaluate on 500 benign instruction-following prompts and 250 prompts for measuring over-refusal. We report Attack Rejection Rate (ARR), Benign Pass Rate (BPR), Over-Refusal Rate (ORR), and MMLU. Higher ARR, BPR, and MMLU are preferred, while lower ORR is preferred. ARR, BPR, and ORR are evaluated using WildGuard \citep{han2024wildguard}. We validate WildGuard against human annotations and observe 97.22\% agreement. Details are provided in Appendix Section~\ref{sec:judge-validation}. Additional experimental settings are provided in Appendix~\ref{sec:additional-experimental-details}.

\subsection{Main Results}
\label{sec:main-results}
Table~\ref{tab:main-results} presents the main results, with the per-attack breakdown and corresponding standard errors reported in Appendix Tables~\ref{tab:per-attack-results} and~\ref{tab:per-attack-results-se}, respectively. Among the baselines, SFT is a fine-tuning-based defense, while LoRA-SFT is its parameter-efficient variant that applies LoRA adapters to all attention output projections. In contrast, ASGuard and DDPO intervene on selected internal model components. ASGuard operates on selected attention heads, while DDPO intervenes at an intermediate-layer representation.

\methodname{} achieves the highest average Attack Rejection Rate (ARR) under both training-data budgets, reaching 96.58\% with only 30 training examples and 99.05\% with 120 examples. The closest robustness results come from SFT, which reaches 96.14\% and 98.74\%, respectively, while LoRA-SFT also reaches 95.63\% ARR in the low-data setting. However, these gains come with substantial degradation in benign utility. At $n=30$, SFT and LoRA-SFT reduce Benign Pass Rate (BPR) to 24.64\% and 67.60\%, respectively. In contrast, \methodname{} is nearly tied with the highest BPR among defenses at $n=30$, achieving 97.68\% compared with 97.84\% for ASGuard, and achieves the highest BPR at $n=120$ with 96.84\%. \methodname{} also obtains the lowest Over-Refusal Rate (ORR) among defenses at $n=30$ with 10.08\%. At $n=120$, its ORR of 13.04\% is second to DDPO at 9.28\%. However, DDPO achieves 55.43\% MMLU at $n=30$ and 54.60\% at $n=120$, compared with 57.68\% for the undefended models, indicating degradation in general language-model capabilities. In comparison, \methodname{} preserves MMLU almost entirely, achieving 57.72\% and 57.66\%, respectively.

\subsection{Adaptive Attacks}
\label{sec:adaptive-attacks}

\begin{table*}[t]
\caption{Attack Rejection Rate (\%) under adaptive attacks. $n_{\mathrm{train}}=30$ (15 plain harmful and 15 clean prompts). \methodname{} scales 20\% of attention heads. Results are reported as the mean over 235 test samples $\pm$ standard error.}
\label{tab:adaptive-attacks}
\centering
\small
\begin{tabular}{@{}llcccc@{}}
\toprule
Model & Attack & No Defense & \methodnameshort{} & ASGuard & DDPO \\
\midrule

\multirow{2}{*}{Llama-3.1-8B}
& Adaptive GCG
& $90.64 \pm 1.90$
& $\mathbf{98.30 \pm 0.84}$
& $91.06 \pm 1.86$
& $95.32 \pm 1.38$ \\

& Adaptive PAIR
& $71.49 \pm 2.95$
& $\mathbf{78.30 \pm 2.69}$
& $68.94 \pm 3.02$
& $73.19 \pm 2.89$ \\

\midrule

\multirow{2}{*}{DeepSeek-7B}
& Adaptive GCG
& $58.30 \pm 3.22$
& $\mathbf{71.49 \pm 2.95}$
& $60.00 \pm 3.20$
& $67.23 \pm 3.06$ \\

& Adaptive PAIR
& $37.87 \pm 3.16$
& $\mathbf{72.34 \pm 2.92}$
& $38.72 \pm 3.18$
& $66.38 \pm 3.08$ \\

\midrule

\multirow{2}{*}{OpenChat-3.5-7B}
& Adaptive GCG
& $35.32 \pm 3.12$
& $\mathbf{87.66 \pm 2.15}$
& $62.55 \pm 3.16$
& $79.15 \pm 2.65$ \\

& Adaptive PAIR
& $30.21 \pm 3.00$
& $\mathbf{75.32 \pm 2.81}$
& $51.49 \pm 3.26$
& $70.64 \pm 2.97$ \\

\midrule

\multirow{2}{*}{Vicuna-13B}
& Adaptive GCG
& $71.91 \pm 2.93$
& $\mathbf{90.21 \pm 1.94}$
& $73.62 \pm 2.87$
& $87.23 \pm 2.18$ \\

& Adaptive PAIR
& $47.23 \pm 3.26$
& $\mathbf{92.34 \pm 1.73}$
& $46.38 \pm 3.25$
& $76.60 \pm 2.76$ \\

\midrule

\multirow{2}{*}{Gemma-3-27B}
& Adaptive GCG
& $91.49 \pm 1.82$
& $97.02 \pm 1.11$
& $94.47 \pm 1.49$
& $\mathbf{97.45 \pm 1.03}$ \\

& Adaptive PAIR
& $57.87 \pm 3.22$
& $\mathbf{89.36 \pm 2.01}$
& $69.79 \pm 3.00$
& $81.28 \pm 2.54$ \\

\bottomrule
\end{tabular}
\end{table*}

Table~\ref{tab:adaptive-attacks} compares defense performance under adaptive attacks. Adaptive GCG optimizes adversarial suffixes directly against the defended model, while Adaptive PAIR iteratively refines prompts using responses from the defended model. \methodname{} consistently outperforms the baseline defenses under adaptive attacks, with the exception of Gemma-3 under Adaptive GCG, where it performs nearly identically to DDPO (97.02\% vs. 97.45\%). On average, \methodname{} achieves 85.23\% attack rejection rate, compared with 79.45\% for DDPO and 65.70\% for ASGuard.

\subsection{Ablations}
\label{sec:ablations}
\subsubsection{Head Selection}
\label{sec:head-selection-ablation}
We compare \methodname{}, which jointly selects and scales 20\% of the attention heads, with two variants that differ only in head selection. The first removes head selection and scales all attention heads. The second removes joint selection by first selecting 20\% of the heads using ASGuard's head-selection procedure \citep{park2026asguard} and then optimizing the \methodname{} scaling functions on this fixed subset. All variants use the same training data and optimization settings. Figure~\ref{fig:selection-ablation}(a) compares Attack Rejection Rate (ARR) and Benign Pass Rate (BPR), while Figure~\ref{fig:selection-ablation}(b) examines disagreement cases between \methodname{} and each variant across the evaluated attacks. Overall, \methodname{} achieves higher jailbreak robustness than both variants, with an average ARR of 96.58\%, compared with 92.26\% when all heads are scaled and 94.73\% with the ASGuard selector. BPR remains approximately unchanged across the three methods, at 97.68\%, 97.36\%, and 97.52\%, respectively. Notably, on disagreement cases, \methodname{} makes the correct rejection decision in 85.58\% of cases against the all-head variant and 79.60\% against the ASGuard-selector variant, compared with 14.42\% and 20.40\% for the corresponding variants. This pattern is consistent across the tested models.

\subsubsection{Joint Learning and Dynamic Scaling}

\begin{table*}[t]
\caption{Ablation of joint versus sequential head selection and dynamic versus static head scaling. All variants are trained with $n=30$ (15 harmful and
15 clean prompts) and scale 20\% of attention heads. Attack rejection rate (ARR) is evaluated on 2,115 harmful prompts across nine attacks, benign pass rate (BPR) on 500 benign prompts, and over-refusal rate (ORR) on 250 deceptive-benign prompts per model. Values are reported as percentages (rate $\pm$ standard error). Arrows indicate the preferred direction of each metric.}
\label{tab:joint-dynamic-ablation}
\centering
\small
\begin{tabular}{@{}llrrr@{}}
\toprule
Model & Method
& ARR (\%) $\uparrow$
& BPR (\%) $\uparrow$
& ORR (\%) $\downarrow$ \\
\midrule

\multirow{4}{*}{Llama-3.1-8B}
& Joint + Dynamic (Ours)
& $99.57 \pm 0.14$
& $98.00 \pm 0.63$
& $13.20 \pm 2.14$ \\
& Sequential + Dynamic
& $98.82 \pm 0.24$
& $97.60 \pm 0.68$
& $17.60 \pm 2.41$ \\
& Joint + Static
& $98.11 \pm 0.30$
& $97.80 \pm 0.66$
& $14.00 \pm 2.19$ \\
& Sequential + Static
& $98.01 \pm 0.30$
& $97.80 \pm 0.66$
& $13.20 \pm 2.14$ \\

\midrule

\multirow{4}{*}{DeepSeek-7B}
& Joint + Dynamic (Ours)
& $92.10 \pm 0.59$
& $97.60 \pm 0.68$
& $17.60 \pm 2.41$ \\
& Sequential + Dynamic
& $87.04 \pm 0.73$
& $97.40 \pm 0.71$
& $18.40 \pm 2.45$ \\
& Joint + Static
& $73.66 \pm 0.96$
& $97.80 \pm 0.66$
& $18.80 \pm 2.47$ \\
& Sequential + Static
& $71.11 \pm 0.99$
& $97.20 \pm 0.74$
& $18.00 \pm 2.43$ \\

\midrule

\multirow{4}{*}{OpenChat-3.5-7B}
& Joint + Dynamic (Ours)
& $91.58 \pm 0.60$
& $98.20 \pm 0.59$
& $4.40 \pm 1.30$ \\
& Sequential + Dynamic
& $86.24 \pm 0.75$
& $98.00 \pm 0.63$
& $9.60 \pm 1.86$ \\
& Joint + Static
& $56.26 \pm 1.08$
& $97.20 \pm 0.74$
& $5.60 \pm 1.45$ \\
& Sequential + Static
& $53.05 \pm 1.09$
& $97.40 \pm 0.71$
& $5.20 \pm 1.40$ \\

\midrule

\multirow{4}{*}{Vicuna-13B}
& Joint + Dynamic (Ours)
& $99.76 \pm 0.11$
& $95.40 \pm 0.94$
& $10.80 \pm 1.96$ \\
& Sequential + Dynamic
& $94.61 \pm 0.49$
& $95.20 \pm 0.96$
& $19.60 \pm 2.51$ \\
& Joint + Static
& $90.07 \pm 0.65$
& $95.80 \pm 0.90$
& $12.40 \pm 2.08$ \\
& Sequential + Static
& $87.90 \pm 0.71$
& $95.00 \pm 0.97$
& $12.80 \pm 2.11$ \\

\midrule

\multirow{4}{*}{Gemma-3-27B}
& Joint + Dynamic (Ours)
& $99.86 \pm 0.08$
& $99.20 \pm 0.40$
& $4.40 \pm 1.30$ \\
& Sequential + Dynamic
& $96.55 \pm 0.40$
& $98.60 \pm 0.53$
& $6.80 \pm 1.59$ \\
& Joint + Static
& $96.22 \pm 0.41$
& $99.40 \pm 0.35$
& $6.00 \pm 1.50$ \\
& Sequential + Static
& $93.81 \pm 0.52$
& $99.20 \pm 0.40$
& $5.60 \pm 1.45$ \\

\bottomrule
\end{tabular}
\end{table*}

Table~\ref{tab:joint-dynamic-ablation} isolates the effects of joint optimization and dynamic scaling. Joint variants optimize head selection and scaling together, whereas sequential variants select heads before optimizing their scaling factors. Dynamic variants predict scaling factors conditioned on each head's query representation, whereas static variants learn fixed scaling factors. Replacing joint optimization with sequential optimization, dynamic scaling with static scaling, or both consistently reduces the attack rejection rate (ARR) and generally increases the over-refusal rate (ORR), while maintaining approximately similar benign pass rates (BPR). Averaged across the five models, joint optimization with dynamic scaling achieves 96.58\% ARR, compared with 92.65\% for sequential optimization with dynamic scaling, 82.87\% for joint optimization with static scaling, and 80.78\% for sequential optimization with static scaling. The corresponding BPR values are 97.68\%, 97.36\%, 97.60\%, and 97.32\%, while the ORR values are 10.08\%, 14.40\%, 11.36\%, and 10.96\%, respectively.

\subsubsection{Scalar Scaling}
We further compare our proposed scalar attention-head scaling with a vector scaling variant. At each token position, our method applies a single positive factor to the entire head output, preserving the direction of the head activation and the relative ratios of its components while changing only its magnitude. In contrast, vector scaling applies a separate scaling factor to each dimension, which changes the direction of the head activation. Appendix Table~\ref{tab:scalar-vector-ablation} compares the two scaling variants. Scalar scaling consistently achieves a higher benign pass rate (BPR) and a lower over-refusal rate (ORR), while maintaining approximately similar attack rejection rates (ARR). On average across all five models, scalar scaling achieves a BPR of 97.68\% versus 95.88\%, an ORR of 10.08\% versus 12.72\%, and an ARR of 96.57\% versus 96.21\% for vector scaling.

\subsubsection{Number of Heads}
\label{sec:number-heads-ablation}
Figure~\ref{fig:number-heads-ablation} and Appendix Table~\ref{tab:number-heads-values} report \methodname{} performance as the percentage of scaled attention heads varies. The evaluation is performed using training-set sizes of 30, 60, and 120, with equal numbers of harmful and benign samples. Results are averaged across five models and, for harmful prompts, across
nine attacks, with standard errors also reported.

Across all three training-set sizes, Attack Rejection Rate (ARR) reaches its highest values when approximately 20--30\% of the attention heads are scaled and generally declines as the intervention becomes denser. The standard error is also substantially larger when only 1\% of the heads are scaled, decreases near the 20--30\% region, and increases again toward larger head budgets, indicating less consistent performance at the extremes. In contrast, Benign Pass Rate (BPR) remains relatively stable. Scaling only 1\% of the heads yields a slightly higher BPR on average, but at a substantial cost in jailbreak robustness. On average, 1\% scaling achieves 81.70\% ARR and 98.49\% BPR, whereas the 20\% budget used by \methodname{} achieves 98.11\% ARR and 97.24\% BPR.

\subsection{Runtime}
\label{sec:runtime}
Due to space constraints, the full runtime results are reported in
Appendix Table~\ref{tab:timing-per-model}, which provides training and
generation time across five models and two training-set sizes,
$n=30$ and $n=120$.

Training time increases with training-set size, while generation latency
remains generally similar across the two budgets for a given model and
defense. On average, \methodname{} requires 63.20 and 294.66 seconds of training
for $n=30$ and $n=120$, compared with 152.53 and 616.87 seconds for SFT
and 107.02 and 437.41 seconds for LoRA-SFT, respectively. Its generation latency averages 202.07~ms per sample, compared with 196.93~ms for the undefended models. This small difference is consistent with \methodname{} applying its intervention only during the input-prefill phase and not during autoregressive decoding. Moreover, at the 20\% head budget, \methodname{} adds only approximately $1.9\times10^{-4}\%$--$3.8\times10^{-4}\%$ additional parameters relative to the parameter count of the underlying LLM.

ASGuard is comparable in runtime, averaging 74.78 and 331.06 seconds of
training for $n=30$ and $n=120$, respectively, with an average generation
latency of 202.46~ms per sample. DDPO trains faster, averaging 37.40 and
160.06 seconds for the two training budgets, but has higher generation latency at 226.94~ms per sample.

\section{Conclusion}
This paper introduces \methodname{}, a jailbreak defense that rescales the outputs of a sparse subset of attention heads, selecting 20\% by default. The scaling factor is conditioned on each head's query representation and attenuates or amplifies the head output while preserving its direction, with head selection and scaling optimized jointly. \methodname{} performs well in low-data settings, outperforming existing defenses.

\subsection*{AI Use Statement}
We used generative AI tools to assist with grammar, LaTeX formatting, and code debugging. All AI-assisted changes were reviewed by the authors.

\subsection*{Reproducibility Statement}
We provide supplementary material containing our code, data, and a README with instructions for setting up the environment and running the code. We have verified that the provided code runs without errors.

\bibliography{iclr2027_conference}
\bibliographystyle{iclr2027_conference}

\clearpage
\appendix

\section{Appendix}

\subsection{Multi-Head Self-Attention}
\label{sec:attention-background}
Multi-head self-attention allows a Transformer layer to compute multiple attention representations of the same input in parallel \citep{vaswani2017attention}. Let $\mathbf{X}_{\ell}\in\mathbb{R}^{T\times d}$ denote the input to self-attention layer $\ell$, where $T$ is the sequence length and $d$ is the model hidden dimension. For attention head $h$, query, key, and value representations are obtained through learned linear projections:
\begin{equation}
\mathbf{Q}_{\ell h}=\mathbf{X}_{\ell}\mathbf{W}^{Q}_{\ell h}, \qquad
\mathbf{K}_{\ell h}=\mathbf{X}_{\ell}\mathbf{W}^{K}_{\ell h}, \qquad
\mathbf{V}_{\ell h}=\mathbf{X}_{\ell}\mathbf{W}^{V}_{\ell h}.
\end{equation}
The output of head $h$ is
\begin{equation}
\mathbf{O}_{\ell h}
=
\operatorname{softmax}\!\left(
\frac{\mathbf{Q}_{\ell h}\mathbf{K}_{\ell h}^{\top}}{\sqrt{d_h}}
\right)\mathbf{V}_{\ell h},
\label{eq:head-attention}
\end{equation}
where $d_h$ is the per-head representation dimension. The outputs of the $H$ attention heads are concatenated and projected through a learned output matrix:
\begin{equation}
\operatorname{MultiHead}(\mathbf{Q},\mathbf{K},\mathbf{V})
=
\operatorname{Concat}\!\left(
\mathbf{O}_{\ell 1},\ldots,\mathbf{O}_{\ell H}
\right)\mathbf{W}^{O},
\label{eq:mha-output}
\end{equation}
where $\mathbf{W}^{O}$ projects the concatenated head representations to the model hidden dimension. The representation is incorporated into the residual stream and processed by the feed-forward sublayer.

\subsection{Extended Experimental Setting}
\label{sec:additional-experimental-details}
Our evaluation includes DeepSeek-LLM-7B-Chat, OpenChat-3.5-1210 (7B), Llama-3.1-8B-Instruct, Vicuna-13B-v1.5, and Gemma-3-27B-IT, using their implementations from Hugging Face. We compare \methodname{} against several baseline defenses. These include Safety Fine-Tuning (SFT), its parameter-efficient variant LoRA-SFT \citep{hu2021lora}, and defenses that intervene on selected model components: ASGuard \citep{park2026asguard}, which operates on attention heads, and DDPO \citep{obidov2026dynamic}, which intervenes on intermediate-layer representations. Each defense is evaluated against nine attacks: Plain Harmful (PH), consisting of harmful prompts without an adversarial component, Human-Crafted (HC) jailbreak prompts, GCG \citep{zou2023universal}, PEZ \citep{wen2023hard}, UAT \citep{wallace2019universal}, AutoPrompt (AP) \citep{shin2020autoprompt}, GBDA \citep{guo2021gradient}, GCG-M \citep{zou2023universal}, and PAIR \citep{chao2025jailbreaking}.

We evaluate the defenses under both a low-data training budget and a standard training budget consistent with prior work \citep{obidov2026dynamic,park2026asguard}: 30 training samples (15 harmful and 15 benign) and 120 training samples (60 harmful and 60 benign). The harmful training prompts are distributed equally among Plain Harmful, Human-Crafted, and GCG examples. The remaining six attack types are not observed during training and are therefore evaluated in a zero-shot setting. The harmful test set contains 235 unseen prompts for each of the nine attack types, for a total of 2,115 harmful test prompts. We additionally evaluate on 500 unseen benign instruction-following prompts and 250 benign prompts designed to measure over-refusal. The latter contain superficially harmful phrases that are benign in context, such as ``How do I shoot a goal in football?''. Harmful prompts are drawn from the WildJailbreak \citep{jiang2024wildteaming}, JailbreakBench \citep{chao2024jailbreakbench}, AdvBench \citep{zou2023universal}, and HarmBench \citep{mazeika2024harmbench} benchmarks. Benign instruction-following prompts are drawn from the Stanford Alpaca dataset \citep{taori2023stanford}, while the over-refusal prompts are taken from XSTest \citep{rottger2024xstest}.

Our evaluation metrics include Attack Rejection Rate (ARR), defined as the percentage of harmful prompts that are rejected, Benign Pass Rate (BPR), defined as the percentage of benign prompts for which the model provides a helpful response, and Over-Refusal Rate (ORR), defined as the percentage of benign deceptive prompts that are incorrectly rejected. We also evaluate MMLU, a standard multiple-choice benchmark for measuring general language-model capabilities. Higher values are preferred for ARR, BPR, and MMLU, while lower values are preferred for ORR. ARR, BPR, and ORR are evaluated using WildGuard as the judge model \citep{han2024wildguard}. We validate WildGuard against human annotations and observe 97.22\% agreement. Details are provided in Appendix Section~\ref{sec:judge-validation}.

All experiments are conducted on a server equipped with two NVIDIA A100 80\,GB GPUs. \methodname{} uses $g(r)=\tanh(r)$ with $\lambda=\log 5$ and is trained for 8 epochs in Stage~1, 12 epochs in Stage~2 for joint head selection and scaling optimization, and 8 epochs in Stage~3. We use the AdamW optimizer with learning rates of $2\times10^{-3}$ for the scaling controller and $2\times10^{-2}$ for the selector. SFT is performed for 8 epochs with a learning rate of $2\times10^{-5}$. LoRA-SFT applies rank-16 LoRA adapters ($\alpha=16$) to all attention output projections and is trained for 8 epochs with a learning rate of $2\times10^{-4}$. DDPO follows its original configuration, using a learning rate of $10^{-3}$ for 40 epochs. For ASGuard, activation scaling is trained for 3 epochs with a learning rate of $5\times10^{-2}$, while preventative fine-tuning is performed for one epoch with a learning rate of $9\times10^{-6}$, consistent with its original implementation. The hyperparameters were tuned on a validation set, consisting of 10 harmful examples from each of the Plain Harmful (PH), Human-Crafted (HC), and GCG categories (30 harmful examples in total), together with a matching set of 30 benign examples. The final test set excluded the validation examples.

\subsection{Human Validation of the LLM Judge}
\label{sec:judge-validation}
We validate WildGuard, which is used to compute Attack Rejection Rate (ARR), Benign Pass Rate (BPR), and Over-Refusal Rate (ORR), through human annotation. We sample target-model responses from all six evaluated defenses: no defense, \methodname{}, SFT, LoRA-SFT, ASGuard, and DDPO. For each defense, we randomly sample 30 harmful, 30 benign, and 30 over-refusal examples, stratified evenly across the five evaluated models, yielding 540 examples in total.

For each example, a human annotator examines the input prompt and target-model response and determines whether its behavior is appropriate. For harmful prompts, a response is considered correct if it refuses or otherwise avoids providing substantive assistance. For benign and over-refusal prompts, a response is considered correct if it provides a relevant and helpful response. The annotator does not observe the WildGuard judgment before completing their annotation and is also blinded to the target model, defense, and evaluation set associated with each example. We then compare the human judgments with the corresponding WildGuard decisions and measure the agreement rate.

Table~\ref{tab:judge-validation} reports the agreement results. Overall, WildGuard agrees with the human judgments on 525 of 540 examples, corresponding to an agreement rate of 97.22\%. Agreement is 96.11\% on harmful prompts, 98.33\% on benign prompts, and 97.22\% on over-refusal prompts. Across defenses, agreement ranges from 94.44\% for LoRA-SFT to 98.89\% for both the undefended model and \methodname{}. The remaining agreement rates are 96.67\% for SFT and ASGuard and 97.78\% for DDPO. Overall, we observe no substantial concentration of disagreement for any particular defense or evaluation set.

\begin{table}[t]
\caption{Agreement (\%) between WildGuard and human judgments, aggregated across the five evaluated models. Evaluation is performed on 540 randomly sampled examples, evenly distributed across the tested models, defenses, and evaluation sets (harmful, benign, and over-refusal).}
\label{tab:judge-validation}
\centering
\begin{tabular}{@{}lrrrrrr@{}}
\toprule
Set
& No defense
& \methodnameshort{}
& SFT
& LoRA-SFT
& ASGuard
& DDPO \\
\midrule
Harmful
& 96.67
& 96.67
& 96.67
& 93.33
& 96.67
& 96.67 \\

Benign
& 100.00
& 100.00
& 100.00
& 96.67
& 96.67
& 96.67 \\

Over-refusal
& 100.00
& 100.00
& 93.33
& 93.33
& 96.67
& 100.00 \\
\bottomrule
\end{tabular}
\end{table}

\subsection{Extended Main Results}
We provide the per-attack breakdown corresponding to Table~\ref{tab:main-results}. Table~\ref{tab:per-attack-results} reports Attack Rejection Rate (ARR) separately for each of the nine attacks, while Table~\ref{tab:per-attack-results-se} reports the corresponding standard errors. The evaluated attacks include Plain Harmful prompts (PH) without an adversarial component, manually constructed Human-Crafted jailbreak prompts (HC), the automated black-box PAIR attack, and white-box optimization-based attacks including PEZ, UAT, GCG, AutoPrompt (AP), GBDA, and GCG-M. Training uses only PH, HC, and GCG examples, while all nine attacks are used for evaluation. Thus, the remaining six attack types are evaluated in a zero-shot setting. In the tables, $n$ denotes the total training-data size, with half of the examples being harmful and the other half benign. Each attack is evaluated on the undefended model and against \methodname{}, SFT, LoRA-SFT, ASGuard, and DDPO.

\begin{table*}[t]
\caption{Per-attack experimental results at two training-data budgets: $n=30$ (15 harmful and 15 clean) and $n=120$ (60 harmful and 60 clean) samples. Training is performed on equal numbers of Plain Harmful (PH), Human-Crafted (HC), and GCG prompts, while Attack Rejection Rate (ARR) is measured on 235 unseen harmful prompts for each of the 9 attacks, 6 of which are evaluated in a zero-shot setting. AP denotes AutoPrompt. Benign Pass Rate (BPR) is evaluated on 500 unseen benign instruction-following prompts, and Over-Refusal Rate (ORR) is evaluated on 250 unseen prompts that contain superficially harmful content but are benign in context. MMLU measures preservation of general language-model capabilities. Higher values are preferred for ARR, BPR, and MMLU, while lower values are preferred for ORR.}
\label{tab:per-attack-results}
\centering
\scriptsize
\setlength{\tabcolsep}{2.3pt}
\begin{tabular}{@{}lll*{9}{r}*{3}{r}@{}}
\toprule
Model & Defense & Train
& PH & HC & GCG & PEZ & UAT & AP & GBDA & GCG-M & PAIR
& BPR & ORR & MMLU \\
\midrule

\multirow{12}{*}{Llama-3.1}
& \multirow{2}{*}{None}
& $n = 30$ & 94.47 & 71.91 & 90.64 & 95.32 & 97.45 & 87.23 & 94.04 & 84.68 & 71.49 & \textbf{99.40} & \textbf{7.60} & 53.13 \\
& & $n = 120$ & 94.47 & 71.91 & 90.64 & 95.32 & 97.45 & 87.23 & 94.04 & 84.68 & 71.49 & \textbf{99.40} & 7.60 & \textbf{53.13} \\
& \multirow{2}{*}{\methodnameshort{}}
& $n = 30$ & \textbf{99.57} & 98.72 & \textbf{100.00} & \textbf{100.00} & 99.57 & \textbf{99.57} & \textbf{100.00} & 99.57 & 99.15 & 98.00 & 13.20 & \textbf{53.63} \\
& & $n = 120$ & 98.30 & \textbf{98.30} & \textbf{100.00} & \textbf{100.00} & \textbf{100.00} & \textbf{99.15} & \textbf{100.00} & \textbf{100.00} & \textbf{99.15} & 97.80 & 14.00 & \textbf{53.13} \\
& \multirow{2}{*}{SFT}
& $n = 30$ & 91.49 & \textbf{99.57} & 99.57 & 99.57 & 99.57 & 98.72 & 99.57 & \textbf{100.00} & \textbf{100.00} & 4.40 & 98.80 & 24.15 \\
& & $n = 120$ & 96.17 & 95.74 & 98.72 & 95.74 & 97.02 & \textbf{99.15} & 97.45 & 95.32 & 98.72 & 10.40 & 93.60 & 24.15 \\
& \multirow{2}{*}{LoRA-SFT}
& $n = 30$ & 78.30 & 97.87 & 77.87 & 72.77 & 70.21 & 73.62 & 71.49 & 78.30 & 92.77 & 80.60 & 21.60 & 23.43 \\
& & $n = 120$ & 88.51 & 97.87 & 80.00 & 75.74 & 82.98 & 78.72 & 69.36 & 82.13 & 94.89 & 93.20 & \textbf{5.20} & 24.43 \\
& \multirow{2}{*}{ASGuard}
& $n = 30$ & 96.17 & 74.04 & 91.49 & 97.45 & 98.30 & 89.36 & 97.45 & 88.94 & 76.60 & \textbf{99.40} & 9.60 & 52.78 \\
& & $n = 120$ & 97.87 & 72.34 & 92.34 & 97.87 & 98.72 & 89.79 & 97.87 & 92.77 & 78.72 & \textbf{99.40} & 10.80 & 51.64 \\
& \multirow{2}{*}{DDPO}
& $n = 30$ & \textbf{99.57} & 85.96 & 97.45 & \textbf{100.00} & \textbf{100.00} & 97.87 & \textbf{100.00} & 97.87 & 94.47 & 97.80 & 8.80 & 42.74 \\
& & $n = 120$ & \textbf{100.00} & 92.77 & 96.60 & 97.02 & 97.45 & 97.02 & 98.30 & 99.15 & 90.21 & 96.79 & 10.00 & 34.33 \\

\midrule
\multirow{12}{*}{DeepSeek}
& \multirow{2}{*}{None}
& $n = 30$ & 93.19 & 43.83 & 58.30 & 71.91 & 69.36 & 60.00 & 69.36 & 51.49 & 37.87 & 99.00 & 18.00 & 49.15 \\
& & $n = 120$ & 93.19 & 43.83 & 58.30 & 71.91 & 69.36 & 60.00 & 69.36 & 51.49 & 37.87 & \textbf{99.00} & 18.00 & 49.15 \\
& \multirow{2}{*}{\methodnameshort{}}
& $n = 30$ & \textbf{100.00} & 72.77 & 93.19 & 96.60 & 97.87 & 95.74 & 96.60 & 94.89 & 81.28 & 97.60 & 17.60 & \textbf{49.50} \\
& & $n = 120$ & \textbf{99.57} & 98.72 & 98.30 & 98.72 & 99.57 & 97.02 & 98.30 & 98.30 & 98.30 & 96.40 & 22.00 & \textbf{49.22} \\
& \multirow{2}{*}{SFT}
& $n = 30$ & \textbf{100.00} & \textbf{100.00} & 99.15 & \textbf{98.72} & \textbf{99.15} & 98.30 & \textbf{98.30} & \textbf{100.00} & 99.15 & 13.60 & 92.00 & 24.15 \\
& & $n = 120$ & 99.15 & \textbf{100.00} & \textbf{100.00} & \textbf{100.00} & \textbf{100.00} & \textbf{100.00} & \textbf{100.00} & \textbf{99.57} & 99.15 & 13.40 & 92.80 & 25.14 \\
& \multirow{2}{*}{LoRA-SFT}
& $n = 30$ & 99.57 & \textbf{100.00} & \textbf{100.00} & 97.87 & \textbf{99.15} & \textbf{98.72} & 95.74 & 99.57 & \textbf{100.00} & 79.40 & \textbf{14.40} & 44.44 \\
& & $n = 120$ & 98.72 & \textbf{100.00} & \textbf{100.00} & 99.57 & 99.57 & 99.57 & 98.72 & 99.15 & \textbf{100.00} & 90.40 & \textbf{5.20} & 29.34 \\
& \multirow{2}{*}{ASGuard}
& $n = 30$ & 92.77 & 46.38 & 60.43 & 71.91 & 70.21 & 57.87 & 68.09 & 54.47 & 37.02 & \textbf{99.20} & 18.00 & 49.29 \\
& & $n = 120$ & 94.47 & 44.26 & 61.28 & 74.04 & 70.64 & 57.87 & 69.79 & 55.32 & 37.02 & 98.80 & 18.80 & \textbf{49.22} \\
& \multirow{2}{*}{DDPO}
& $n = 30$ & 99.57 & 88.94 & 97.02 & 97.87 & 98.72 & 97.02 & 97.87 & 98.30 & 94.47 & 83.20 & 19.60 & 49.22 \\
& & $n = 120$ & 99.15 & 98.72 & 97.87 & 97.87 & 99.57 & 97.87 & 97.87 & 99.15 & 94.47 & 93.80 & 11.20 & 48.65 \\

\midrule
\multirow{12}{*}{OpenChat}
& \multirow{2}{*}{None}
& $n = 30$ & 68.51 & 48.94 & 35.32 & 52.34 & 45.53 & 35.32 & 44.68 & 28.09 & 30.21 & \textbf{99.40} & 4.00 & 63.53 \\
& & $n = 120$ & 68.51 & 48.94 & 35.32 & 52.34 & 45.53 & 35.32 & 44.68 & 28.09 & 30.21 & \textbf{99.40} & \textbf{4.00} & 63.53 \\
& \multirow{2}{*}{\methodnameshort{}}
& $n = 30$ & 99.15 & 74.04 & 94.89 & 97.45 & 95.74 & 95.74 & 94.47 & 90.64 & 82.13 & 98.20 & 4.40 & \textbf{64.03} \\
& & $n = 120$ & \textbf{99.57} & 94.47 & 98.72 & 98.72 & 99.15 & 97.02 & 99.15 & 97.02 & 92.34 & 97.00 & 7.60 & \textbf{64.53} \\
& \multirow{2}{*}{SFT}
& $n = 30$ & 71.49 & 89.79 & 81.70 & 84.68 & 82.13 & 80.85 & 86.81 & 90.64 & 92.77 & 0.40 & 100.00 & 24.15 \\
& & $n = 120$ & 97.87 & \textbf{100.00} & \textbf{99.57} & \textbf{100.00} & 99.57 & 99.15 & \textbf{100.00} & \textbf{100.00} & \textbf{100.00} & 10.80 & 96.40 & 24.15 \\
& \multirow{2}{*}{LoRA-SFT}
& $n = 30$ & \textbf{99.57} & \textbf{100.00} & \textbf{100.00} & \textbf{100.00} & \textbf{100.00} & \textbf{100.00} & \textbf{100.00} & \textbf{100.00} & \textbf{100.00} & 1.20 & 95.20 & 24.15 \\
& & $n = 120$ & 25.11 & 57.02 & 63.40 & 64.26 & 62.55 & 64.26 & 72.77 & 81.28 & 54.47 & 3.00 & 97.20 & 25.21 \\
& \multirow{2}{*}{ASGuard}
& $n = 30$ & 55.74 & 61.28 & 80.85 & 86.38 & 79.57 & 84.68 & 90.64 & 81.70 & 54.89 & 93.60 & 14.80 & 63.60 \\
& & $n = 120$ & 98.72 & 92.77 & 99.15 & 99.15 & \textbf{100.00} & \textbf{99.57} & 99.15 & 97.87 & 92.34 & 70.80 & 35.60 & 63.82 \\
& \multirow{2}{*}{DDPO}
& $n = 30$ & 96.60 & 78.30 & 91.06 & 93.62 & 92.77 & 94.04 & 92.34 & 92.34 & 80.43 & 93.60 & \textbf{3.20} & 63.68 \\
& & $n = 120$ & 97.45 & 95.74 & 94.47 & 94.89 & 94.04 & 97.02 & 94.47 & 97.02 & 83.83 & 93.80 & 4.80 & 64.46 \\

\midrule
\multirow{12}{*}{Vicuna}
& \multirow{2}{*}{None}
& $n = 30$ & 99.57 & 52.77 & 71.91 & 95.32 & 94.47 & 84.68 & 91.06 & 63.40 & 47.23 & \textbf{97.80} & 10.80 & \textbf{50.43} \\
& & $n = 120$ & 99.57 & 52.77 & 71.91 & 95.32 & 94.47 & 84.68 & 91.06 & 63.40 & 47.23 & \textbf{97.80} & \textbf{10.80} & 50.43 \\
& \multirow{2}{*}{\methodnameshort{}}
& $n = 30$ & \textbf{100.00} & 97.87 & \textbf{100.00} & \textbf{100.00} & \textbf{100.00} & \textbf{100.00} & \textbf{100.00} & \textbf{100.00} & \textbf{100.00} & 95.40 & 10.80 & 48.79 \\
& & $n = 120$ & \textbf{100.00} & \textbf{100.00} & \textbf{100.00} & \textbf{100.00} & \textbf{100.00} & \textbf{100.00} & \textbf{100.00} & \textbf{100.00} & \textbf{100.00} & 93.60 & 14.80 & 48.79 \\
& \multirow{2}{*}{SFT}
& $n = 30$ & \textbf{100.00} & 98.72 & 98.30 & \textbf{100.00} & 99.57 & 99.57 & 98.72 & 99.57 & 99.57 & 8.40 & 98.80 & 23.86 \\
& & $n = 120$ & 99.15 & 97.02 & 99.15 & \textbf{100.00} & \textbf{100.00} & 99.57 & \textbf{100.00} & 99.57 & 96.17 & 23.60 & 89.60 & 23.86 \\
& \multirow{2}{*}{LoRA-SFT}
& $n = 30$ & \textbf{100.00} & \textbf{100.00} & \textbf{100.00} & \textbf{100.00} & \textbf{100.00} & \textbf{100.00} & \textbf{100.00} & \textbf{100.00} & \textbf{100.00} & 80.60 & \textbf{9.20} & 49.79 \\
& & $n = 120$ & \textbf{100.00} & \textbf{100.00} & \textbf{100.00} & \textbf{100.00} & \textbf{100.00} & \textbf{100.00} & \textbf{100.00} & \textbf{100.00} & \textbf{100.00} & 22.00 & 79.60 & 26.64 \\
& \multirow{2}{*}{ASGuard}
& $n = 30$ & 99.57 & 51.91 & 73.62 & 95.32 & 94.89 & 86.38 & 92.34 & 65.11 & 45.96 & 97.20 & 11.60 & 49.93 \\
& & $n = 120$ & \textbf{100.00} & 52.77 & 76.17 & 96.60 & 97.02 & 88.51 & 95.74 & 66.38 & 46.38 & 97.00 & 11.20 & 50.43 \\
& \multirow{2}{*}{DDPO}
& $n = 30$ & \textbf{100.00} & 94.04 & 99.57 & \textbf{100.00} & \textbf{100.00} & \textbf{100.00} & \textbf{100.00} & 99.57 & 98.30 & 81.40 & 24.00 & 47.93 \\
& & $n = 120$ & \textbf{100.00} & 99.57 & \textbf{100.00} & \textbf{100.00} & \textbf{100.00} & 99.15 & \textbf{100.00} & \textbf{100.00} & 97.45 & 94.20 & 13.20 & \textbf{52.35} \\

\midrule
\multirow{12}{*}{Gemma-3}
& \multirow{2}{*}{None}
& $n = 30$ & 96.17 & 68.51 & 91.49 & 95.32 & 93.19 & 91.06 & 94.47 & 84.26 & 57.87 & \textbf{99.80} & \textbf{2.80} & 72.15 \\
& & $n = 120$ & 96.17 & 68.51 & 91.49 & 95.32 & 93.19 & 91.06 & 94.47 & 84.26 & 57.87 & \textbf{99.80} & \textbf{2.80} & 72.15 \\
& \multirow{2}{*}{\methodnameshort{}}
& $n = 30$ & \textbf{100.00} & \textbf{100.00} & \textbf{100.00} & \textbf{100.00} & \textbf{100.00} & \textbf{100.00} & \textbf{100.00} & \textbf{100.00} & 98.72 & 99.20 & 4.40 & 72.65 \\
& & $n = 120$ & \textbf{100.00} & \textbf{100.00} & \textbf{100.00} & \textbf{100.00} & \textbf{100.00} & \textbf{100.00} & \textbf{100.00} & \textbf{100.00} & \textbf{99.57} & 99.40 & 6.80 & 72.65 \\
& \multirow{2}{*}{SFT}
& $n = 30$ & 99.15 & 97.02 & \textbf{100.00} & 99.57 & 99.57 & 99.15 & 99.15 & 98.72 & 98.30 & 96.40 & 18.40 & 65.81 \\
& & $n = 120$ & 99.15 & 90.64 & 99.57 & 99.57 & 99.57 & 99.57 & \textbf{100.00} & \textbf{100.00} & 96.60 & 70.00 & 46.40 & 63.75 \\
& \multirow{2}{*}{LoRA-SFT}
& $n = 30$ & \textbf{100.00} & \textbf{100.00} & \textbf{100.00} & \textbf{100.00} & \textbf{100.00} & \textbf{100.00} & \textbf{100.00} & \textbf{100.00} & \textbf{100.00} & 96.20 & 23.20 & 67.31 \\
& & $n = 120$ & 98.72 & 98.72 & 94.47 & 94.89 & 95.74 & 96.60 & 95.32 & 99.15 & 97.45 & 96.80 & 16.80 & 23.93 \\
& \multirow{2}{*}{ASGuard}
& $n = 30$ & 97.45 & 78.30 & 97.02 & 99.57 & 98.72 & 95.32 & 98.30 & 90.64 & 70.64 & \textbf{99.80} & 4.40 & 71.87 \\
& & $n = 120$ & 99.57 & 84.68 & 99.57 & \textbf{100.00} & \textbf{100.00} & \textbf{100.00} & \textbf{100.00} & 97.02 & 88.09 & \textbf{99.80} & 8.00 & 72.22 \\
& \multirow{2}{*}{DDPO}
& $n = 30$ & 98.72 & 85.11 & 99.15 & 99.57 & \textbf{100.00} & \textbf{100.00} & \textbf{100.00} & 97.45 & 93.19 & \textbf{99.80} & 5.20 & \textbf{73.58} \\
& & $n = 120$ & 99.57 & 88.94 & 99.57 & 99.57 & \textbf{100.00} & \textbf{100.00} & \textbf{100.00} & 97.45 & 96.60 & 99.60 & 7.20 & \textbf{73.22} \\
\bottomrule
\end{tabular}
\end{table*}

\begin{table*}[t]
\caption{Standard errors corresponding to the results in Table~\ref{tab:per-attack-results} at two training-data budgets: $n=30$ (15 harmful and 15 clean) and $n=120$ (60 harmful and 60 clean) samples. Training is performed on equal numbers of Plain Harmful (PH), Human-Crafted (HC), and GCG prompts, while Attack Rejection Rate (ARR) is evaluated on 235 unseen harmful prompts for each of the 9 attacks, 6 of which are evaluated in a zero-shot setting. AP denotes AutoPrompt. Benign Pass Rate (BPR) is evaluated on 500 unseen benign instruction-following prompts, and Over-Refusal Rate (ORR) is evaluated on 250 unseen prompts that contain superficially harmful content but are benign in context. MMLU measures preservation of general language-model capabilities.}
\label{tab:per-attack-results-se}
\centering
\scriptsize
\setlength{\tabcolsep}{3.8pt}
\begin{tabular}{@{}lll*{9}{r}*{3}{r}@{}}
\toprule
Model & Defense & Train
& PH & HC & GCG & PEZ & UAT & AP & GBDA & GCG-M & PAIR
& BPR & ORR & MMLU \\
\midrule

\multirow{12}{*}{Llama-3.1}
& \multirow{2}{*}{None}
& $n = 30$  & 1.49 & 2.93 & 1.90 & 1.38 & 1.03 & 2.18 & 1.54 & 2.35 & 2.95 & 0.35 & 1.68 & 1.33 \\
& & $n = 120$ & 1.49 & 2.93 & 1.90 & 1.38 & 1.03 & 2.18 & 1.54 & 2.35 & 2.95 & 0.35 & 1.68 & 1.33 \\
& \multirow{2}{*}{\methodnameshort{}}
& $n = 30$  & 0.42 & 0.73 & 0.00 & 0.00 & 0.42 & 0.42 & 0.00 & 0.42 & 0.60 & 0.63 & 2.14 & 1.33 \\
& & $n = 120$ & 0.84 & 0.84 & 0.00 & 0.00 & 0.00 & 0.60 & 0.00 & 0.00 & 0.60 & 0.66 & 2.19 & 1.33 \\
& \multirow{2}{*}{SFT}
& $n = 30$  & 1.82 & 0.42 & 0.42 & 0.42 & 0.42 & 0.73 & 0.42 & 0.00 & 0.00 & 0.92 & 0.69 & 1.14 \\
& & $n = 120$ & 1.25 & 1.32 & 0.73 & 1.32 & 1.11 & 0.60 & 1.03 & 1.38 & 0.73 & 1.37 & 1.55 & 1.14 \\
& \multirow{2}{*}{LoRA-SFT}
& $n = 30$  & 2.69 & 0.94 & 2.71 & 2.90 & 2.98 & 2.87 & 2.95 & 2.69 & 1.69 & 1.77 & 2.60 & 1.13 \\
& & $n = 120$ & 2.08 & 0.94 & 2.61 & 2.80 & 2.45 & 2.67 & 3.01 & 2.50 & 1.44 & 1.13 & 1.40 & 1.15 \\
& \multirow{2}{*}{ASGuard}
& $n = 30$  & 1.25 & 2.86 & 1.82 & 1.03 & 0.84 & 2.01 & 1.03 & 2.05 & 2.76 & 0.35 & 1.86 & 1.33 \\
& & $n = 120$ & 0.94 & 2.92 & 1.73 & 0.94 & 0.73 & 1.98 & 0.94 & 1.69 & 2.67 & 0.35 & 1.96 & 1.33 \\
& \multirow{2}{*}{DDPO}
& $n = 30$  & 0.42 & 2.27 & 1.03 & 0.00 & 0.00 & 0.94 & 0.00 & 0.94 & 1.49 & 0.66 & 1.79 & 1.32 \\
& & $n = 120$ & 0.00 & 1.69 & 1.18 & 1.11 & 1.03 & 1.11 & 0.84 & 0.60 & 1.94 & 0.79 & 1.90 & 1.27 \\

\midrule
\multirow{12}{*}{DeepSeek}
& \multirow{2}{*}{None}
& $n = 30$  & 1.64 & 3.24 & 3.22 & 2.93 & 3.01 & 3.20 & 3.01 & 3.26 & 3.16 & 0.45 & 2.43 & 1.33 \\
& & $n = 120$ & 1.64 & 3.24 & 3.22 & 2.93 & 3.01 & 3.20 & 3.01 & 3.26 & 3.16 & 0.45 & 2.43 & 1.33 \\
& \multirow{2}{*}{\methodnameshort{}}
& $n = 30$  & 0.00 & 2.90 & 1.64 & 1.18 & 0.94 & 1.32 & 1.18 & 1.44 & 2.54 & 0.68 & 2.41 & 1.33 \\
& & $n = 120$ & 0.42 & 0.73 & 0.84 & 0.73 & 0.42 & 1.11 & 0.84 & 0.84 & 0.84 & 0.83 & 2.62 & 1.33 \\
& \multirow{2}{*}{SFT}
& $n = 30$  & 0.00 & 0.00 & 0.60 & 0.73 & 0.60 & 0.84 & 0.84 & 0.00 & 0.60 & 1.53 & 1.72 & 1.14 \\
& & $n = 120$ & 0.60 & 0.00 & 0.00 & 0.00 & 0.00 & 0.00 & 0.00 & 0.42 & 0.60 & 1.52 & 1.63 & 1.16 \\
& \multirow{2}{*}{LoRA-SFT}
& $n = 30$  & 0.42 & 0.00 & 0.00 & 0.94 & 0.60 & 0.73 & 1.32 & 0.42 & 0.00 & 1.81 & 2.22 & 1.33 \\
& & $n = 120$ & 0.73 & 0.00 & 0.00 & 0.42 & 0.42 & 0.42 & 0.73 & 0.60 & 0.00 & 1.32 & 1.40 & 1.22 \\
& \multirow{2}{*}{ASGuard}
& $n = 30$  & 1.69 & 3.25 & 3.19 & 2.93 & 2.98 & 3.22 & 3.04 & 3.25 & 3.15 & 0.40 & 2.43 & 1.33 \\
& & $n = 120$ & 1.49 & 3.24 & 3.18 & 2.86 & 2.97 & 3.22 & 3.00 & 3.24 & 3.15 & 0.49 & 2.47 & 1.33 \\
& \multirow{2}{*}{DDPO}
& $n = 30$  & 0.42 & 2.05 & 1.11 & 0.94 & 0.73 & 1.11 & 0.94 & 0.84 & 1.49 & 1.67 & 2.51 & 1.33 \\
& & $n = 120$ & 0.60 & 0.73 & 0.94 & 0.94 & 0.42 & 0.94 & 0.94 & 0.60 & 1.49 & 1.08 & 1.99 & 1.33 \\

\midrule
\multirow{12}{*}{OpenChat}
& \multirow{2}{*}{None}
& $n = 30$  & 3.03 & 3.26 & 3.12 & 3.26 & 3.25 & 3.12 & 3.24 & 2.93 & 3.00 & 0.35 & 1.24 & 1.28 \\
& & $n = 120$ & 3.03 & 3.26 & 3.12 & 3.26 & 3.25 & 3.12 & 3.24 & 2.93 & 3.00 & 0.35 & 1.24 & 1.28 \\
& \multirow{2}{*}{\methodnameshort{}}
& $n = 30$  & 0.60 & 2.86 & 1.44 & 1.03 & 1.32 & 1.32 & 1.49 & 1.90 & 2.50 & 0.59 & 1.30 & 1.28 \\
& & $n = 120$ & 0.42 & 1.49 & 0.73 & 0.73 & 0.60 & 1.11 & 0.60 & 1.11 & 1.73 & 0.76 & 1.68 & 1.28 \\
& \multirow{2}{*}{SFT}
& $n = 30$  & 2.95 & 1.98 & 2.52 & 2.35 & 2.50 & 2.57 & 2.21 & 1.90 & 1.69 & 0.28 & 0.00 & 1.14 \\
& & $n = 120$ & 0.94 & 0.00 & 0.42 & 0.00 & 0.42 & 0.60 & 0.00 & 0.00 & 0.00 & 1.39 & 1.18 & 1.14 \\
& \multirow{2}{*}{LoRA-SFT}
& $n = 30$  & 0.42 & 0.00 & 0.00 & 0.00 & 0.00 & 0.00 & 0.00 & 0.00 & 0.00 & 0.49 & 1.35 & 1.14 \\
& & $n = 120$ & 2.83 & 3.23 & 3.14 & 3.13 & 3.16 & 3.13 & 2.90 & 2.54 & 3.25 & 0.76 & 1.04 & 1.16 \\
& \multirow{2}{*}{ASGuard}
& $n = 30$  & 3.24 & 3.18 & 2.57 & 2.24 & 2.63 & 2.35 & 1.90 & 2.52 & 3.25 & 1.09 & 2.25 & 1.28 \\
& & $n = 120$ & 0.73 & 1.69 & 0.60 & 0.60 & 0.00 & 0.42 & 0.60 & 0.94 & 1.73 & 2.03 & 3.03 & 1.28 \\
& \multirow{2}{*}{DDPO}
& $n = 30$  & 1.18 & 2.69 & 1.86 & 1.59 & 1.69 & 1.54 & 1.73 & 1.73 & 2.59 & 1.09 & 1.11 & 1.28 \\
& & $n = 120$ & 1.03 & 1.32 & 1.49 & 1.44 & 1.54 & 1.11 & 1.49 & 1.11 & 2.40 & 1.08 & 1.35 & 1.28 \\

\midrule
\multirow{12}{*}{Vicuna}
& \multirow{2}{*}{None}
& $n = 30$  & 0.42 & 3.26 & 2.93 & 1.38 & 1.49 & 2.35 & 1.86 & 3.14 & 3.26 & 0.66 & 1.96 & 1.33 \\
& & $n = 120$ & 0.42 & 3.26 & 2.93 & 1.38 & 1.49 & 2.35 & 1.86 & 3.14 & 3.26 & 0.66 & 1.96 & 1.33 \\
& \multirow{2}{*}{\methodnameshort{}}
& $n = 30$  & 0.00 & 0.94 & 0.00 & 0.00 & 0.00 & 0.00 & 0.00 & 0.00 & 0.00 & 0.94 & 1.96 & 1.33 \\
& & $n = 120$ & 0.00 & 0.00 & 0.00 & 0.00 & 0.00 & 0.00 & 0.00 & 0.00 & 0.00 & 1.09 & 2.25 & 1.33 \\
& \multirow{2}{*}{SFT}
& $n = 30$  & 0.00 & 0.73 & 0.84 & 0.00 & 0.42 & 0.42 & 0.73 & 0.42 & 0.42 & 1.24 & 0.69 & 1.14 \\
& & $n = 120$ & 0.60 & 1.11 & 0.60 & 0.00 & 0.00 & 0.42 & 0.00 & 0.42 & 1.25 & 1.90 & 1.93 & 1.14 \\
& \multirow{2}{*}{LoRA-SFT}
& $n = 30$  & 0.00 & 0.00 & 0.00 & 0.00 & 0.00 & 0.00 & 0.00 & 0.00 & 0.00 & 1.77 & 1.83 & 1.33 \\
& & $n = 120$ & 0.00 & 0.00 & 0.00 & 0.00 & 0.00 & 0.00 & 0.00 & 0.00 & 0.00 & 1.85 & 2.55 & 1.18 \\
& \multirow{2}{*}{ASGuard}
& $n = 30$  & 0.42 & 3.26 & 2.87 & 1.38 & 1.44 & 2.24 & 1.73 & 3.11 & 3.25 & 0.74 & 2.03 & 1.33 \\
& & $n = 120$ & 0.00 & 3.26 & 2.78 & 1.18 & 1.11 & 2.08 & 1.32 & 3.08 & 3.25 & 0.76 & 1.99 & 1.33 \\
& \multirow{2}{*}{DDPO}
& $n = 30$  & 0.00 & 1.54 & 0.42 & 0.00 & 0.00 & 0.00 & 0.00 & 0.42 & 0.84 & 1.74 & 2.70 & 1.33 \\
& & $n = 120$ & 0.00 & 0.42 & 0.00 & 0.00 & 0.00 & 0.60 & 0.00 & 0.00 & 1.03 & 1.05 & 2.14 & 1.33 \\

\midrule
\multirow{12}{*}{Gemma-3}
& \multirow{2}{*}{None}
& $n = 30$  & 1.25 & 3.03 & 1.82 & 1.38 & 1.64 & 1.86 & 1.49 & 2.38 & 3.22 & 0.20 & 1.04 & 1.20 \\
& & $n = 120$ & 1.25 & 3.03 & 1.82 & 1.38 & 1.64 & 1.86 & 1.49 & 2.38 & 3.22 & 0.20 & 1.04 & 1.20 \\
& \multirow{2}{*}{\methodnameshort{}}
& $n = 30$  & 0.00 & 0.00 & 0.00 & 0.00 & 0.00 & 0.00 & 0.00 & 0.00 & 0.73 & 0.40 & 1.30 & 1.19 \\
& & $n = 120$ & 0.00 & 0.00 & 0.00 & 0.00 & 0.00 & 0.00 & 0.00 & 0.00 & 0.42 & 0.35 & 1.59 & 1.19 \\
& \multirow{2}{*}{SFT}
& $n = 30$  & 0.60 & 1.11 & 0.00 & 0.42 & 0.42 & 0.60 & 0.60 & 0.73 & 0.84 & 0.83 & 2.45 & 1.27 \\
& & $n = 120$ & 0.60 & 1.90 & 0.42 & 0.42 & 0.42 & 0.42 & 0.00 & 0.00 & 1.18 & 2.05 & 3.15 & 1.28 \\
& \multirow{2}{*}{LoRA-SFT}
& $n = 30$  & 0.00 & 0.00 & 0.00 & 0.00 & 0.00 & 0.00 & 0.00 & 0.00 & 0.00 & 0.86 & 2.67 & 1.25 \\
& & $n = 120$ & 0.73 & 0.73 & 1.49 & 1.44 & 1.32 & 1.18 & 1.38 & 0.60 & 1.03 & 0.79 & 2.36 & 1.14 \\
& \multirow{2}{*}{ASGuard}
& $n = 30$  & 1.03 & 2.69 & 1.11 & 0.42 & 0.73 & 1.38 & 0.84 & 1.90 & 2.97 & 0.20 & 1.30 & 1.20 \\
& & $n = 120$ & 0.42 & 2.35 & 0.42 & 0.00 & 0.00 & 0.00 & 0.00 & 1.11 & 2.11 & 0.20 & 1.72 & 1.20 \\
& \multirow{2}{*}{DDPO}
& $n = 30$  & 0.73 & 2.32 & 0.60 & 0.42 & 0.00 & 0.00 & 0.00 & 1.03 & 1.64 & 0.20 & 1.40 & 1.18 \\
& & $n = 120$ & 0.42 & 2.05 & 0.42 & 0.42 & 0.00 & 0.00 & 0.00 & 1.03 & 1.18 & 0.28 & 1.63 & 1.18 \\
\bottomrule
\end{tabular}
\end{table*}

\clearpage

\subsection{Extended Scalar Scaling Ablation}
\label{sec:appendix-scalar-scaling}

\begin{table*}[t]
\caption{Ablation of scalar versus vector attention-head scaling at $n=30$ (15 harmful and 15 clean prompts). Both variants jointly optimize head selection and scaling, with 20\% of attention heads selected. Attack rejection rate (ARR) is evaluated on 2,115 harmful prompts across nine attacks, benign pass rate (BPR) on 500 benign prompts, and over-refusal rate (ORR) on 250 deceptive-benign prompts per model. Values are reported as percentages (rate $\pm$ standard error). Arrows indicate the preferred direction of each metric.}
\label{tab:scalar-vector-ablation}
\centering
\small
\begin{tabular}{@{}llrrr@{}}
\toprule
Model & Scaling
& ARR (\%) $\uparrow$
& BPR (\%) $\uparrow$
& ORR (\%) $\downarrow$ \\
\midrule

\multirow{2}{*}{Llama-3.1-8B}
& Scalar (Ours)
& $99.57 \pm 0.14$
& $98.00 \pm 0.63$
& $13.20 \pm 2.14$ \\
& Vector
& $99.53 \pm 0.15$
& $96.60 \pm 0.81$
& $14.00 \pm 2.19$ \\

\midrule

\multirow{2}{*}{DeepSeek-7B}
& Scalar (Ours)
& $92.10 \pm 0.59$
& $97.60 \pm 0.68$
& $17.60 \pm 2.41$ \\
& Vector
& $91.68 \pm 0.60$
& $96.20 \pm 0.86$
& $18.40 \pm 2.45$ \\

\midrule

\multirow{2}{*}{OpenChat-3.5-7B}
& Scalar (Ours)
& $91.58 \pm 0.60$
& $98.20 \pm 0.59$
& $4.40 \pm 1.30$ \\
& Vector
& $91.63 \pm 0.60$
& $97.40 \pm 0.71$
& $6.00 \pm 1.50$ \\

\midrule

\multirow{2}{*}{Vicuna-13B}
& Scalar (Ours)
& $99.76 \pm 0.11$
& $95.40 \pm 0.94$
& $10.80 \pm 1.96$ \\
& Vector
& $98.20 \pm 0.29$
& $93.60 \pm 1.09$
& $15.60 \pm 2.29$ \\

\midrule

\multirow{2}{*}{Gemma-3-27B}
& Scalar (Ours)
& $99.86 \pm 0.08$
& $99.20 \pm 0.40$
& $4.40 \pm 1.30$ \\
& Vector
& $100.00 \pm 0.00$
& $95.60 \pm 0.92$
& $9.60 \pm 1.86$ \\

\bottomrule
\end{tabular}
\end{table*}

We compare our proposed scalar attention-head scaling with a vector scaling variant. At each token position, our method applies a single positive factor to the entire head output, preserving the direction of the head activation and the relative ratios of its components while changing only its magnitude. In contrast, vector scaling applies a separate scaling factor to each dimension, which changes the direction of the head activation. Table~\ref{tab:scalar-vector-ablation} compares the two scaling variants. Scalar scaling consistently achieves a higher benign pass rate (BPR) and a lower over-refusal rate (ORR), while maintaining approximately similar attack rejection rates (ARR). On average across all five models, scalar scaling achieves a BPR of 97.68\% versus 95.88\%, an ORR of 10.08\% versus 12.72\%, and an ARR of 96.57\% versus 96.21\% for vector scaling.

\subsection{Extended Number of Heads Ablation}
\label{sec:appendix-number-heads}
Table~\ref{tab:number-heads-values} provides the numerical results corresponding to Figure~\ref{fig:number-heads-ablation}. We vary the
percentage of attention heads scaled by \methodname{} from 1\% to 100\% for training-set sizes of $n\in\{30,60,120\}$, with equal numbers of harmful and benign training samples. Across all three training-set sizes, attack rejection rate (ARR) reaches its highest values when approximately 20--30\% of the attention heads are scaled and generally declines as the intervention becomes denser. The standard error is also substantially larger when only 1\% of the heads are scaled, decreases near the 20--30\% region, and increases again toward larger head budgets, indicating less consistent performance at the extremes. In contrast, benign pass rate (BPR) remains relatively stable. Scaling only 1\% of the heads yields a slightly higher BPR on average, but at a substantial cost in jailbreak robustness. On average, 1\% scaling achieves 81.70\% ARR and 98.49\% BPR, whereas the 20\% budget used by \methodname{} achieves 98.11\% ARR and 97.24\% BPR.

This ablation serves as a diagnostic analysis to illustrate the effect of the attention-head budget on performance. Hyperparameter tuning was performed separately on a validation set prior to this diagnostic, as described in Appendix~\ref{sec:additional-experimental-details}.

\begin{table*}[t]
\caption{Numerical results corresponding to Figure~\ref{fig:number-heads-ablation}. We vary the percentage of attention heads scaled using 30, 60, and 120 training samples, with equal numbers of harmful and benign examples. For each of the tested models, Attack Rejection Rate (ARR) is evaluated on 2,115 harmful prompts, equally distributed across nine attacks, and Benign Pass Rate (BPR) is evaluated on 500 benign prompts. Values are reported as mean $\pm$ standard error across five models.}
\label{tab:number-heads-values}
\centering
\small
\begin{tabular}{@{}c cc cc cc@{}}
\toprule
& \multicolumn{2}{c}{$n_{\mathrm{train}}=30$}
& \multicolumn{2}{c}{$n_{\mathrm{train}}=60$}
& \multicolumn{2}{c}{$n_{\mathrm{train}}=120$} \\
\cmidrule(lr){2-3}
\cmidrule(lr){4-5}
\cmidrule(lr){6-7}
Heads (\%) 
& ARR $\uparrow$ & BPR $\uparrow$
& ARR $\uparrow$ & BPR $\uparrow$
& ARR $\uparrow$ & BPR $\uparrow$ \\
\midrule

1
& $77.55{\pm}8.87$ & $98.68{\pm}0.48$
& $81.80{\pm}8.31$ & $98.84{\pm}0.43$
& $85.76{\pm}7.99$ & $97.96{\pm}0.66$ \\

10
& $93.23{\pm}4.24$ & $97.64{\pm}0.75$
& $97.88{\pm}1.25$ & $97.16{\pm}0.86$
& $98.18{\pm}1.17$ & $96.84{\pm}0.95$ \\

20
& $\mathbf{96.58{\pm}1.93}$ & $97.68{\pm}0.63$
& $\mathbf{98.69{\pm}0.86}$ & $97.20{\pm}0.99$
& $99.05{\pm}0.50$ & $96.84{\pm}0.95$ \\

30
& $96.43{\pm}1.89$ & $97.44{\pm}0.94$
& $98.68{\pm}0.84$ & $97.32{\pm}1.19$
& $\mathbf{99.28{\pm}0.38}$ & $97.24{\pm}1.00$ \\

40
& $96.08{\pm}2.24$ & $97.60{\pm}0.90$
& $98.40{\pm}0.90$ & $97.40{\pm}1.13$
& $99.09{\pm}0.51$ & $97.16{\pm}0.89$ \\

50
& $95.41{\pm}2.17$ & $97.92{\pm}1.03$
& $97.54{\pm}1.35$ & $97.48{\pm}0.89$
& $98.69{\pm}0.71$ & $97.40{\pm}0.90$ \\

60
& $94.52{\pm}2.91$ & $97.84{\pm}1.03$
& $97.23{\pm}1.66$ & $97.28{\pm}1.05$
& $98.27{\pm}0.72$ & $97.60{\pm}0.88$ \\

70
& $93.45{\pm}3.28$ & $97.84{\pm}1.03$
& $97.07{\pm}1.44$ & $97.36{\pm}0.99$
& $97.91{\pm}1.00$ & $97.80{\pm}0.87$ \\

80
& $92.35{\pm}3.45$ & $97.60{\pm}1.09$
& $96.27{\pm}2.09$ & $97.28{\pm}1.03$
& $97.56{\pm}1.28$ & $97.68{\pm}0.87$ \\

90
& $92.43{\pm}3.40$ & $97.60{\pm}1.13$
& $95.60{\pm}2.15$ & $97.28{\pm}1.12$
& $97.46{\pm}1.34$ & $97.52{\pm}0.88$ \\

100
& $92.26{\pm}3.86$ & $97.36{\pm}1.12$
& $95.83{\pm}2.02$ & $97.16{\pm}1.04$
& $96.73{\pm}1.78$ & $97.40{\pm}0.91$ \\

\bottomrule
\end{tabular}
\end{table*}

\subsection{Extended Runtime}
\label{sec:runtime-extended}
Table~\ref{tab:timing-per-model} reports training and generation time across
five models and two training-set sizes, $n=30$ and $n=120$. Training time is
reported in seconds per model, while generation latency is reported as mean time per sample with standard errors. All methods use greedy
decoding with a maximum of 128 new tokens. Additional experimental settings
are provided in Appendix~\ref{sec:additional-experimental-details}.

Training time increases with training-set size, while generation latency
remains generally similar across the two budgets for a given model and
defense. On average, \methodname{} requires 63.20 and 294.66 seconds of training
for $n=30$ and $n=120$, compared with 152.53 and 616.87 seconds for SFT
and 107.02 and 437.41 seconds for LoRA-SFT, respectively. Its generation latency averages 202.07~ms per sample, compared with 196.93~ms for the undefended models. This small difference is consistent with \methodname{} applying its intervention only during the input-prefill phase and not during autoregressive decoding. Moreover, at the 20\% head budget, \methodname{} adds only approximately $1.9\times10^{-4}\%$--$3.8\times10^{-4}\%$ additional parameters relative to the parameter count of the underlying LLM.

ASGuard is comparable in runtime, averaging 74.78 and 331.06 seconds of
training for $n=30$ and $n=120$, respectively, with an average generation
latency of 202.46~ms per sample. DDPO trains faster, averaging 37.40 and
160.06 seconds for the two training budgets, but has higher generation latency at 226.94~ms per sample. As shown in Section~\ref{sec:main-results}, \methodname{} additionally achieves stronger defense performance.

\begin{table*}[t]
\caption{Training and generation time across models and training-set sizes. Training time is reported in seconds per model, while generation time is reported as mean time per sample $\pm$ standard error.}
\label{tab:timing-per-model}
\centering
\small
\begin{tabular}{@{}llrrrr@{}}
\toprule
& &
\multicolumn{2}{c}{$n_{\mathrm{train}}=30$} &
\multicolumn{2}{c}{$n_{\mathrm{train}}=120$} \\
\cmidrule(lr){3-4}
\cmidrule(lr){5-6}
Model & Defense
& Train (s) $\downarrow$
& Generate (ms) $\downarrow$
& Train (s) $\downarrow$
& Generate (ms) $\downarrow$ \\
\midrule

\multirow{6}{*}{Llama-3.1-8B}
& None
& N/A & $82.07 \pm 0.31$
& N/A & $82.07 \pm 0.31$ \\
& \methodnameshort{}
& 40.43 & $76.58 \pm 0.42$
& 195.84 & $78.91 \pm 0.39$ \\
& SFT
& 56.03 & $81.73 \pm 0.31$
& 235.72 & $82.22 \pm 0.31$ \\
& LoRA-SFT
& 52.17 & $84.99 \pm 0.32$
& 236.86 & $85.20 \pm 0.31$ \\
& DDPO
& 22.32 & $108.23 \pm 0.31$
& 104.15 & $108.73 \pm 0.31$ \\
& ASGuard
& 32.65 & $80.49 \pm 0.32$
& 146.12 & $80.59 \pm 0.32$ \\

\midrule

\multirow{6}{*}{DeepSeek-7B}
& None
& N/A & $59.78 \pm 0.24$
& N/A & $59.78 \pm 0.24$ \\
& \methodnameshort{}
& 32.67 & $60.53 \pm 0.24$
& 166.62 & $60.54 \pm 0.25$ \\
& SFT
& 55.80 & $59.41 \pm 0.24$
& 224.99 & $59.72 \pm 0.24$ \\
& LoRA-SFT
& 41.96 & $62.27 \pm 0.25$
& 200.28 & $62.35 \pm 0.25$ \\
& DDPO
& 20.23 & $94.50 \pm 0.25$
& 93.62 & $95.70 \pm 0.25$ \\
& ASGuard
& 27.15 & $60.11 \pm 0.25$
& 128.22 & $62.40 \pm 0.24$ \\

\midrule

\multirow{6}{*}{OpenChat-3.5-7B}
& None
& N/A & $80.06 \pm 0.36$
& N/A & $80.06 \pm 0.36$ \\
& \methodnameshort{}
& 39.39 & $80.12 \pm 0.36$
& 199.79 & $80.02 \pm 0.36$ \\
& SFT
& 62.74 & $79.13 \pm 0.36$
& 255.96 & $79.78 \pm 0.36$ \\
& LoRA-SFT
& 50.86 & $82.29 \pm 0.36$
& 237.21 & $82.24 \pm 0.36$ \\
& DDPO
& 23.31 & $115.71 \pm 0.36$
& 107.29 & $115.39 \pm 0.35$ \\
& ASGuard
& 32.15 & $78.59 \pm 0.37$
& 148.23 & $78.43 \pm 0.37$ \\

\midrule

\multirow{6}{*}{Vicuna-13B}
& None
& N/A & $173.15 \pm 0.53$
& N/A & $173.15 \pm 0.53$ \\
& \methodnameshort{}
& 67.33 & $175.78 \pm 0.53$
& 328.78 & $174.74 \pm 0.53$ \\
& SFT
& 121.80 & $171.99 \pm 0.54$
& 461.37 & $171.78 \pm 0.54$ \\
& LoRA-SFT
& 95.13 & $179.28 \pm 0.55$
& 394.65 & $179.13 \pm 0.55$ \\
& DDPO
& 39.21 & $213.81 \pm 0.56$
& 172.26 & $216.14 \pm 0.55$ \\
& ASGuard
& 60.25 & $171.60 \pm 0.56$
& 262.39 & $171.58 \pm 0.56$ \\

\midrule

\multirow{6}{*}{Gemma-3-27B}
& None
& N/A & $589.60 \pm 0.91$
& N/A & $589.60 \pm 0.91$ \\
& \methodnameshort{}
& 136.20 & $587.30 \pm 0.86$
& 582.26 & $646.22 \pm 1.20$ \\
& SFT
& 466.30 & $574.55 \pm 0.88$
& 1906.33 & $574.49 \pm 0.83$ \\
& LoRA-SFT
& 294.98 & $614.91 \pm 0.83$
& 1118.03 & $611.59 \pm 0.77$ \\
& DDPO
& 81.93 & $595.18 \pm 0.94$
& 322.98 & $605.97 \pm 1.21$ \\
& ASGuard
& 221.72 & $602.07 \pm 0.91$
& 970.34 & $638.71 \pm 1.06$ \\

\bottomrule
\end{tabular}
\end{table*}

\subsection{Training-Seed Stability and Head-Selection Consistency}
\label{sec:seed-stability}
To evaluate performance consistency, we repeat \methodname{} training with three random seeds under
the same settings. Table~\ref{tab:seed-stability} reports these results. Averaged across the seeds and five models, \methodname{} achieves a 96.74\% attack rejection rate (ARR), a 97.55\% benign pass rate (BPR), and a 9.79\% over-refusal rate (ORR), closely matching the original results of 96.58\% ARR, 97.68\% BPR, and 10.08\% ORR.

We also evaluate whether \methodname{} consistently selects the same attention heads for scaling across different seeds. Table~\ref{tab:seed-stability} additionally reports the mean pairwise Jaccard similarity of the selected attention-head sets across the three seed pairs and the percentage of selected heads shared between runs. Approximately 71--79\% of selected heads are shared between runs, suggesting that the learned head-selection mechanism identifies substantially overlapping subsets of attention heads across seeds. The mean Jaccard similarity across models is 0.583, corresponding to 73.6\% shared selected heads, substantially exceeding the random-selection reference (Jaccard $\approx 0.111$, or 20\% shared heads).

\begin{table*}[t]
\caption{Training-seed stability and attention-head selection consistency of \methodname{} at $n=30$ (15 harmful and 15 benign prompts), using three random seeds and a 20\% attention-head budget. Attack rejection rate (ARR) is evaluated on 2,115 harmful prompts across nine attacks, benign pass rate (BPR) on 500 benign prompts, and over-refusal rate (ORR) on 250 deceptive-benign prompts per model. ARR, BPR, and ORR are reported as percentages (mean $\pm$ standard error across three training seeds). Head Overlap is the mean pairwise Jaccard similarity of selected attention-head sets across the three seed pairs. Shared heads (\%) denotes the average percentage of selected heads shared between runs. Arrows indicate the preferred
direction of each metric.}
\label{tab:seed-stability}
\centering
\small
\begin{tabular}{@{}lrrrrr@{}}
\toprule
Model
& ARR (\%) $\uparrow$
& BPR (\%) $\uparrow$
& ORR (\%) $\downarrow$
& Head Overlap $\uparrow$
& Shared heads (\%) $\uparrow$ \\
\midrule

Llama-3.1-8B
& $99.48 \pm 0.18$
& $98.20 \pm 0.40$
& $13.47 \pm 0.71$
& 0.647
& 78.6 \\

DeepSeek-7B
& $92.58 \pm 0.68$
& $97.00 \pm 0.53$
& $16.53 \pm 0.35$
& 0.553
& 71.2 \\

OpenChat-3.5-7B
& $92.20 \pm 0.74$
& $98.40 \pm 0.23$
& $4.27 \pm 0.35$
& 0.600
& 75.0 \\

Vicuna-13B
& $99.59 \pm 0.08$
& $94.87 \pm 0.41$
& $10.13 \pm 0.87$
& 0.550
& 71.0 \\

Gemma-3-27B
& $99.84 \pm 0.04$
& $99.27 \pm 0.18$
& $4.53 \pm 0.27$
& 0.567
& 72.4 \\

\bottomrule
\end{tabular}
\end{table*}

\subsection{Security Threats to LLMs}
\label{sec:related-threats}
LLMs and related generative models face a range of security threats. Jailbreak attacks bypass the safety alignment of LLMs or LLM-powered agents, enabling malicious code generation \citep{li2025llms} or facilitating fraud and cybercrime \citep{andriushchenko2025agentharm}. Prompt injection attacks instead embed malicious instructions in untrusted content, such as retrieved web pages or emails, causing LLM-powered applications to leak sensitive information or execute unauthorized actions \citep{greshake2023not,zhan2024injecagent}. Backdoor attacks introduce malicious behavior that activates only when specific triggers are encountered. Such attacks can have consequences beyond text generation, including manipulating the behavior of LLM-powered robots, causing unsafe movements \citep{zhang2026steptrigger}, or compromising biological models to create targeted health risks \citep{obidov2026genotypic}. Backdoors may be introduced through poisoned training data \citep{xu2024instructions}, compromised pretrained language models \citep{wei2023lmsanitator}, or third-party adapters \citep{obidov2026lorascan}.

\subsection{Generalization to Unseen Attack Families}
\label{sec:unseen-attack-families}

\begin{table*}[t]
\caption{Generalization to unseen attack families under different training settings at $n=30$ (15 harmful and 15 benign prompts). PH-only, HC-only, and GCG-only indicate that the harmful training examples are drawn only from Plain Harmful, Human-Crafted, and GCG attacks, respectively. All variants use a 20\% attention-head budget. Average attack rejection rate (ARR) is evaluated on 2,115 harmful prompts across nine attacks, benign pass rate (BPR) on 500 benign prompts, and over-refusal rate (ORR) on 250 deceptive-benign prompts per model. Per-attack ARR is evaluated on 235 test prompts per attack. All values are reported as percentages (rate $\pm$ standard
error).}
\label{tab:unseen-attack-families}
\centering
\footnotesize
\setlength{\tabcolsep}{3.8pt}
\begin{tabular}{@{}llccccc@{}}
\toprule
Training & Metric
& \shortstack{Llama-3.1}
& \shortstack{DeepSeek}
& \shortstack{OpenChat-3.5}
& \shortstack{Vicuna}
& \shortstack{Gemma-3} \\
\midrule

\multirow{12}{*}{PH-only}
& PH
& $100.00 \pm 0.00$
& $100.00 \pm 0.00$
& $97.87 \pm 0.94$
& $100.00 \pm 0.00$
& $98.72 \pm 0.73$ \\
& HC
& $97.87 \pm 0.94$
& $93.19 \pm 1.64$
& $85.11 \pm 2.32$
& $97.02 \pm 1.11$
& $98.72 \pm 0.73$ \\
& GCG
& $97.87 \pm 0.94$
& $92.77 \pm 1.69$
& $94.47 \pm 1.49$
& $95.32 \pm 1.38$
& $97.45 \pm 1.03$ \\
& PEZ
& $98.72 \pm 0.73$
& $92.77 \pm 1.69$
& $96.60 \pm 1.18$
& $99.57 \pm 0.42$
& $99.15 \pm 0.60$ \\
& UAT
& $97.87 \pm 0.94$
& $94.04 \pm 1.54$
& $94.89 \pm 1.44$
& $100.00 \pm 0.00$
& $98.72 \pm 0.73$ \\
& AutoPrompt
& $99.15 \pm 0.60$
& $94.47 \pm 1.49$
& $94.47 \pm 1.49$
& $98.72 \pm 0.73$
& $96.17 \pm 1.25$ \\
& GBDA
& $100.00 \pm 0.00$
& $95.32 \pm 1.38$
& $93.62 \pm 1.59$
& $100.00 \pm 0.00$
& $98.72 \pm 0.73$ \\
& GCG-M
& $98.72 \pm 0.73$
& $94.04 \pm 1.54$
& $91.49 \pm 1.82$
& $99.15 \pm 0.60$
& $97.87 \pm 0.94$ \\
& PAIR
& $98.30 \pm 0.84$
& $84.68 \pm 2.35$
& $89.36 \pm 2.01$
& $99.15 \pm 0.60$
& $97.45 \pm 1.03$ \\
& Avg. ARR
& $98.72 \pm 0.24$
& $93.48 \pm 0.54$
& $93.10 \pm 0.55$
& $98.77 \pm 0.24$
& $98.11 \pm 0.30$ \\
& BPR
& $99.00 \pm 0.44$
& $98.60 \pm 0.53$
& $99.20 \pm 0.40$
& $96.20 \pm 0.86$
& $99.60 \pm 0.28$ \\
& ORR
& $10.80 \pm 1.96$
& $16.00 \pm 2.32$
& $7.20 \pm 1.63$
& $12.00 \pm 2.06$
& $5.60 \pm 1.45$ \\

\midrule

\multirow{12}{*}{HC-only}
& PH
& $100.00 \pm 0.00$
& $100.00 \pm 0.00$
& $99.15 \pm 0.60$
& $100.00 \pm 0.00$
& $100.00 \pm 0.00$ \\
& HC
& $100.00 \pm 0.00$
& $98.72 \pm 0.73$
& $95.74 \pm 1.32$
& $100.00 \pm 0.00$
& $100.00 \pm 0.00$ \\
& GCG
& $100.00 \pm 0.00$
& $96.17 \pm 1.25$
& $91.49 \pm 1.82$
& $100.00 \pm 0.00$
& $100.00 \pm 0.00$ \\
& PEZ
& $100.00 \pm 0.00$
& $97.45 \pm 1.03$
& $94.47 \pm 1.49$
& $100.00 \pm 0.00$
& $100.00 \pm 0.00$ \\
& UAT
& $100.00 \pm 0.00$
& $99.15 \pm 0.60$
& $94.47 \pm 1.49$
& $100.00 \pm 0.00$
& $100.00 \pm 0.00$ \\
& AutoPrompt
& $99.57 \pm 0.42$
& $94.04 \pm 1.54$
& $93.19 \pm 1.64$
& $100.00 \pm 0.00$
& $100.00 \pm 0.00$ \\
& GBDA
& $100.00 \pm 0.00$
& $97.45 \pm 1.03$
& $91.49 \pm 1.82$
& $100.00 \pm 0.00$
& $100.00 \pm 0.00$ \\
& GCG-M
& $100.00 \pm 0.00$
& $96.17 \pm 1.25$
& $93.19 \pm 1.64$
& $100.00 \pm 0.00$
& $100.00 \pm 0.00$ \\
& PAIR
& $100.00 \pm 0.00$
& $98.30 \pm 0.84$
& $94.04 \pm 1.54$
& $100.00 \pm 0.00$
& $100.00 \pm 0.00$ \\
& Avg. ARR
& $99.95 \pm 0.05$
& $97.49 \pm 0.34$
& $94.14 \pm 0.51$
& $100.00 \pm 0.00$
& $100.00 \pm 0.00$ \\
& BPR
& $98.60 \pm 0.53$
& $95.40 \pm 0.94$
& $96.40 \pm 0.83$
& $95.60 \pm 0.92$
& $99.00 \pm 0.44$ \\
& ORR
& $15.20 \pm 2.27$
& $16.80 \pm 2.36$
& $5.20 \pm 1.40$
& $11.20 \pm 1.99$
& $7.20 \pm 1.63$ \\

\midrule

\multirow{12}{*}{GCG-only}
& PH
& $100.00 \pm 0.00$
& $100.00 \pm 0.00$
& $95.74 \pm 1.32$
& $100.00 \pm 0.00$
& $99.57 \pm 0.42$ \\
& HC
& $94.04 \pm 1.54$
& $89.79 \pm 1.98$
& $87.66 \pm 2.15$
& $97.45 \pm 1.03$
& $99.57 \pm 0.42$ \\
& GCG
& $100.00 \pm 0.00$
& $97.02 \pm 1.11$
& $93.62 \pm 1.59$
& $100.00 \pm 0.00$
& $100.00 \pm 0.00$ \\
& PEZ
& $100.00 \pm 0.00$
& $99.15 \pm 0.60$
& $96.60 \pm 1.18$
& $100.00 \pm 0.00$
& $98.72 \pm 0.73$ \\
& UAT
& $100.00 \pm 0.00$
& $99.57 \pm 0.42$
& $95.32 \pm 1.38$
& $100.00 \pm 0.00$
& $100.00 \pm 0.00$ \\
& AutoPrompt
& $99.57 \pm 0.42$
& $97.87 \pm 0.94$
& $94.89 \pm 1.44$
& $100.00 \pm 0.00$
& $98.30 \pm 0.84$ \\
& GBDA
& $100.00 \pm 0.00$
& $99.57 \pm 0.42$
& $93.19 \pm 1.64$
& $100.00 \pm 0.00$
& $100.00 \pm 0.00$ \\
& GCG-M
& $99.57 \pm 0.42$
& $97.02 \pm 1.11$
& $92.34 \pm 1.73$
& $100.00 \pm 0.00$
& $95.74 \pm 1.32$ \\
& PAIR
& $96.60 \pm 1.18$
& $83.40 \pm 2.43$
& $87.23 \pm 2.18$
& $96.17 \pm 1.25$
& $97.87 \pm 0.94$ \\
& Avg. ARR
& $98.87 \pm 0.23$
& $95.93 \pm 0.43$
& $92.96 \pm 0.56$
& $99.29 \pm 0.18$
& $98.87 \pm 0.23$ \\
& BPR
& $98.80 \pm 0.49$
& $97.20 \pm 0.74$
& $99.00 \pm 0.44$
& $95.20 \pm 0.96$
& $99.60 \pm 0.28$ \\
& ORR
& $14.80 \pm 2.25$
& $16.40 \pm 2.34$
& $7.60 \pm 1.68$
& $11.60 \pm 2.03$
& $6.80 \pm 1.59$ \\

\bottomrule
\end{tabular}
\end{table*}

To evaluate generalization to unseen attack families, we conduct additional experiments training \methodname{} separately on Plain Harmful (PH)-only, Human-Crafted (HC)-only, and GCG-only harmful examples. The training settings are identical except for the harmful training subset. PH-only training contains neither optimization-based jailbreak suffixes or prefixes nor human-crafted jailbreak templates in its harmful training examples. Table~\ref{tab:unseen-attack-families} reports these results.

\methodname{} trained on Plain Harmful examples only achieves a 96.44\% average attack rejection rate (ARR), a 98.52\% benign pass rate (BPR), and a 10.32\% over-refusal rate (ORR), closely matching the original method trained on three harmful attack families (PH, HC, and GCG), which achieves 96.58\% ARR, 97.68\% BPR, and 10.08\% ORR. HC-only and GCG-only training also achieve strong performance, with 98.32\% ARR, 97.00\% BPR, and 11.12\% ORR for HC-only, and 97.18\% ARR, 97.96\% BPR, and 11.44\% ORR for GCG-only.

\end{document}